\documentclass[times,twocolumn,final,round,authoryear]{elsarticle}

\usepackage{medima}
\usepackage{framed,multirow}
\usepackage{marvosym}
\usepackage{amssymb}
\usepackage{latexsym}
\usepackage{booktabs}
\usepackage{url}
\usepackage{xcolor}
\usepackage{multirow}
\usepackage{comment}
\usepackage{hyperref}
\usepackage{algorithm}
\usepackage{algorithmic}
\usepackage{amsmath}
\usepackage{appendix}
\usepackage{tcolorbox}
\usepackage{makecell}
\newlength\myindent
\definecolor{newcolor}{rgb}{.8,.349,.1}

\journal{Medical Image Analysis}

\begin{document}

\verso{Z. Zhong \textit{et~al.}}

\begin{frontmatter}

\title{Abn-BLIP: Abnormality-aligned Bootstrapping Language-Image Pre-training for Pulmonary Embolism Diagnosis and Report Generation from CTPA}
\author[2,3]{Zhusi Zhong}
\author[4]{Yuli Wang}
\author[2,3]{Lulu Bi}
\author[2,3]{Zhuoqi Ma}
\author[2,3]{Sun Ho Ahn}
\author[3]{Christopher J. Mullin}
\author[6]{Colin F. Greineder}
\author[2,3]{Michael K. Atalay}
\author[2,3]{Scott Collins}
\author[2,3]{Grayson L. Baird}
\author[5]{Cheng Ting Lin}
\author[4]{Webster Stayman} % 8
\author[9]{Todd M. Kolb}
\author[7]{Ihab Kamel}
\author[7]{Harrison X. Bai}
\author[2,3]{Zhicheng Jiao \corref{cor1}}
\cortext[cor1]{Corresponding author.\\ E-mail: ${zhicheng\_jiao@brown.edu}$ (Z. Jiao). }
\address[2]{Department of Diagnostic Imaging, Brown University Health, Providence 02903, USA}
\address[3]{Warren Alpert Medical School of Brown University, Providence 02903, USA}
\address[4]{Department of Biomedical Engineering, Johns Hopkins University School of Medicine, Baltimore 21205, USA}
\address[5]{Department of Radiology and Radiological Sciences, Johns Hopkins University School of Medicine, Baltimore 21205, USA}
\address[6]{Department of Emergency Medicine and Department of Pharmacology, University of Michigan, Ann Arbor 48109, USA}
\address[9]{Johns Hopkins University Division of Pulmonary and Critical Care Medicine, Baltimore 21205, USA}
\address[7]{Department of Radiology, University of Colorado School of Medicine, Aurora 80045, USA}
% \address[8]{Biomedical Engineering Department, Johns Hopkins University, Baltimore 21205, USA}

\begin{abstract}
Medical imaging plays a pivotal role in modern healthcare, with computed tomography pulmonary angiography (CTPA) being a critical tool for diagnosing pulmonary embolism and other thoracic conditions. However, the complexity of interpreting CTPA scans and generating accurate radiology reports remains a significant challenge. This paper introduces Abn-BLIP (Abnormality-aligned Bootstrapping Language-Image Pretraining), an advanced diagnosis model designed to align abnormal findings to generate the accuracy and comprehensiveness of radiology reports. By leveraging learnable queries and cross-modal attention mechanisms, our model demonstrates superior performance in detecting abnormalities, reducing missed findings, and generating structured reports compared to existing methods. Our experiments show that Abn-BLIP outperforms state-of-the-art medical vision-language models and 3D report generation methods in both accuracy and clinical relevance. These results highlight the potential of integrating multimodal learning strategies for improving radiology reporting. The source code is available at https://github.com/zzs95/abn-blip.
\end{abstract}
\end{frontmatter}

%\linenumbers

\begin{figure}
\includegraphics[width=0.485\textwidth]{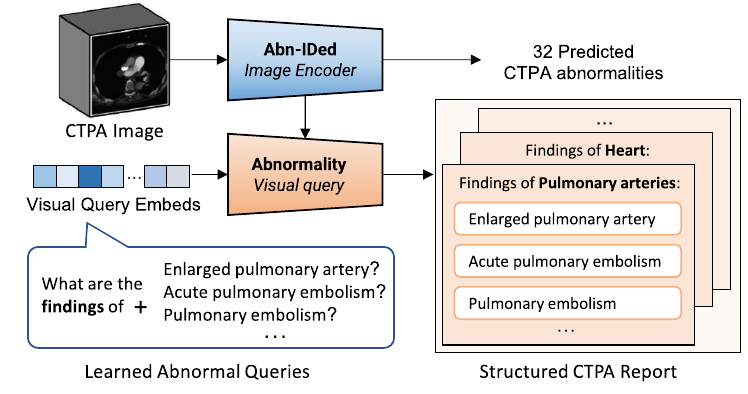}

\caption{Abn-BLIP inference pipeline for CTPA abnormality identification and structured report generation.
The Abn-IDed image encoder detects 32 CTPA abnormalities and extracts abnormality-identified features. The learned visual queries interrogate CTPA scans by Abn-QFormer to extract the corresponding abnormal findings. These queries help generate a structured CTPA report, categorizing abnormalities under relevant organ-specific sections, such as pulmonary arteries and the heart.\label{intro}}
\end{figure}

\section{Introduction}

Pulmonary embolism (PE) is a life-threatening condition caused by thromboembolic obstruction of the pulmonary arteries, often leading to severe complications, long-term morbidity, and high
mortality risk~\citep{buelohlavek2013pulmonary}. Timely and accurate diagnosis is essential for effective treatment and improved
outcomes~\citep{alonso2010delay,hendriksen2017clinical,cahan2023multimodal}. Computed tomography pulmonary angiography (CTPA) remains the reference standard owing to its high sensitivity and
specificity~\citep{CTPA}. However, its interpretation is labor-intensive, reader-dependent, and prone to delays in high-volume clinical environments~\citep{singh2011evaluation}.

Recent advances in medical image AI have demonstrated considerable promise in enhancing PE diagnosis on CTPA~\citep{soffer2021deep}. Deep learning--based multimodal approaches have been developed
to automate embolus identification, quantify clot burden, and stratify patient risk~\citep{huang2020penet,liu2020evaluation,zhong2025pulmonary}, thereby improving efficiency and reducing
inter-reader variability. Nonetheless, most existing systems primarily generate probabilistic predictions with limited interpretability, constraining their clinical
reliability~\citep{huang2020penet,huangMultimodalFusionDeep2020b,lindenmeyer2024inadequacy}. Efforts to incorporate vascular spatial structure~\citep{tajbakhsh2019computer} or artery segmentation
with threshold-based analysis~\citep{pu2023automated} have improved embolus characterization, yet current solutions remain largely restricted to embolism detection rather than providing
comprehensive CTPA assessment, including cardiac function, clot distribution, and ancillary thoracic findings.

Vision--language models (VLMs) represent a promising direction for comprehensive CTPA-based PE assessment by integrating imaging with textual descriptions, thereby enhancing interpretability and
decision support~\citep{wu2025vision,zhong2025vision}. Medical VLMs bridge AI-generated outputs with radiologists' workflows, facilitate automated structured report generation, and reduce
inter-observer variability~\citep{nazi2024large,hartsock2024vision,tanno2024collaboration,jin2024promptmrg}. By incorporating multimodal information, such as clinical scores and patient history,
VLMs enable holistic evaluation for improved patient management and risk stratification~\citep{zhong2024multi}. Unlike conventional models limited to classification or segmentation, VLMs generate
comprehensive, human-readable reports directly from imaging data, thereby enhancing transparency and clinical adoption~\citep{wu2023towards,bai2024m3d,huang2023inspect}.

Despite these advantages, general-purpose medical VLMs remain suboptimal for CTPA-based PE assessment~\citep{hager2024evaluation,zhong2025vision}. Trained on heterogeneous datasets spanning
diverse modalities, these models often lack domain-specific expertise, resulting in reduced sensitivity to subtle radiological findings critical for PE diagnosis. Their ability to address complex
reports and multi-abnormality queries is also limited, constraining integration of visual, textual, and clinical information at a level comparable to expert
radiologists~\citep{hartsock2024vision}. The key challenge is to develop a PE-specific VLM that combines high diagnostic accuracy with interpretability, alignment with radiologists' reporting
conventions, and effective multimodal integration for comprehensive decision support.

To address this gap, we propose Abnormality-aligned Bootstrapping Language--Image Pretraining (Abn-BLIP), a PE-specific VLM that integrates abnormality recognition with structured descriptions for CTPA report generation (Fig. \ref{intro}). Abn-BLIP structures abnormality-specific visual queries into organized diagnostic findings, enabling a multi-stage workflow that enhances interpretability, systematically organizes assessments, and improves the clinical utility of AI-generated radiology reports.
The key contributions are presented as follows:

\begin{itemize}
\item We propose a multi-label abnormality recognition module to enhance diagnostic accuracy in CTPA report generation, with a particular focus on hierarchical analysis of the pulmonary artery region.

\item We introduce Abn-QFormer, which leverages abnormality-driven queries to aggregate image--text features at the abnormality level, dynamically refining cross-modal retrieval and enabling clinician-like examination of individual findings.

\item We develop Abnormality-aligned Contrastive Learning (ACL) to achieve fine-grained alignment between radiological features and textual findings, thereby reinforcing abnormality-level correspondence.

\item Guided by medical diagnostic principles, our framework explicitly models hierarchical relationships between anatomical regions and abnormalities, ensuring comprehensive, structured, and clinically meaningful CTPA reporting.
\end{itemize}

\begin{figure*}[!t]
\centering
\includegraphics[width=0.8\textwidth]{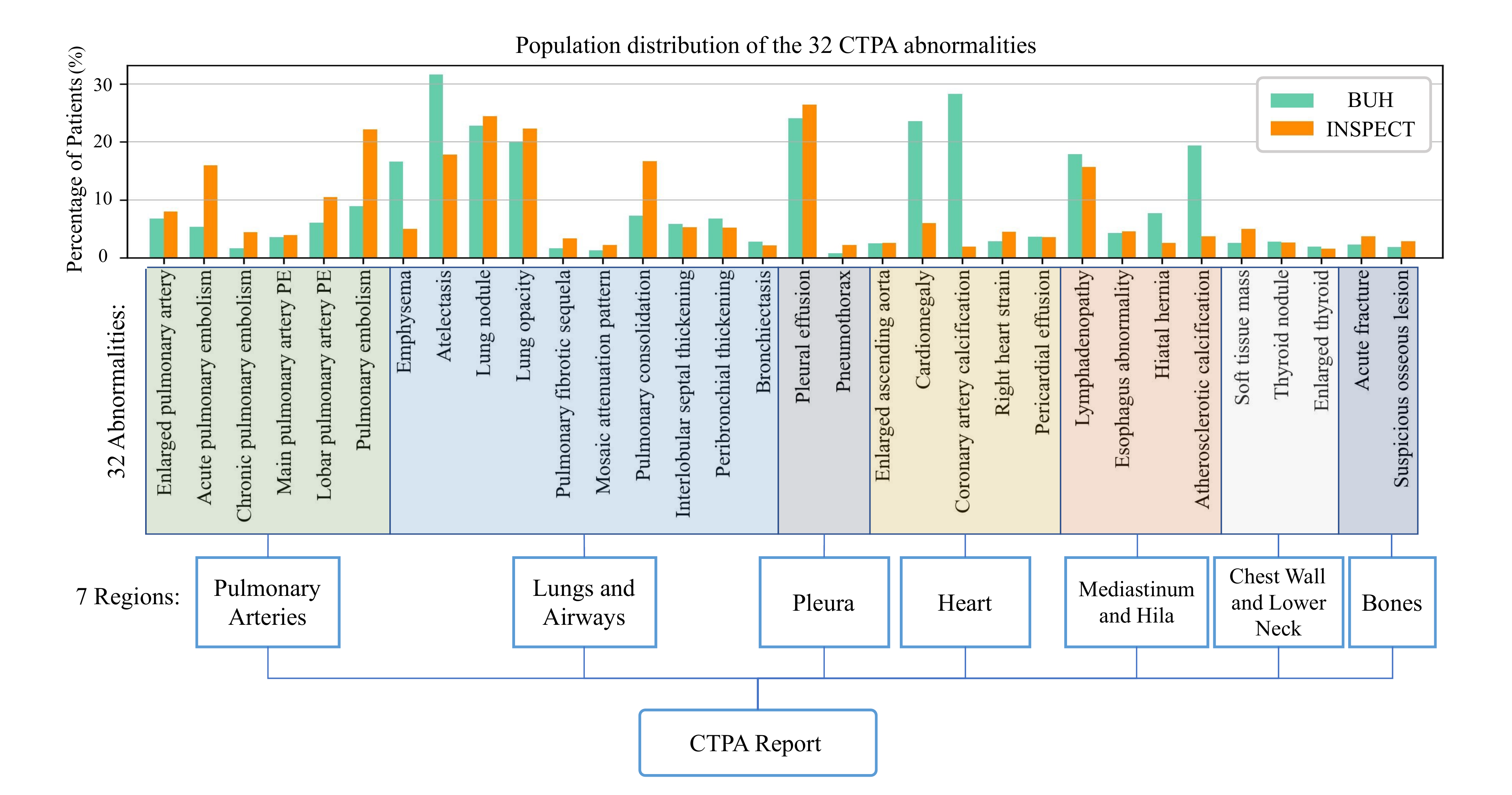}
\caption{The figure illustrates the population distribution of 32 CTPA abnormalities across two datasets (BUH and INSPECT), categorized into 7 anatomical regions: Pulmonary Arteries, Lungs and Airways, Pleura, Heart, Mediastinum and Hila, Chest Wall and Lower Neck, and Bones. This hierarchical framework facilitates comprehensive abnormality detection and enhances the generation of clinically meaningful CTPA reports. The abnormality labels were extracted from radiology reports using a large language model (LLM), enabling a multi-dimensional assessment of inter-regional variations across the datasets.}
\label{fig::abn_bars}
\end{figure*}

\section{Related work}

\textbf{Image Captioning.}
Early image captioning methods predominantly adopted an encoder--decoder framework, in which convolutional neural networks (CNNs) extracted global visual features and recurrent neural networks
(RNNs) generated sequential text~\citep{vinyals2015show,chen2015microsoft}. The introduction of attention mechanisms substantially improved performance by adaptively weighting salient image
regions, thereby overcoming the limitations of fixed-length representations~\citep{xu2015show,lu2017knowing}. Transformer-based architectures~\citep{vaswani2017attention} have been employed to
capture long-range dependencies and support parallel decoding, resulting in more fluent and contextually consistent descriptions.

\textbf{Vision-Language Pre-training.}
Contrastive learning has become a foundational technique in Vision-Language Pre-training, aligning paired images and text within a shared embedding space to enable generalization across tasks.
Models such as CLIP~\citep{radford2021learning} and SimCLR~\citep{khosla2020supervised} optimize InfoNCE losses~\citep{oord2018representation} to learn discriminative representations. Despite
impressive zero-shot capabilities, these approaches often neglect intra-modal structures and lack mechanisms for aligning localized visual semantics with fine-grained textual descriptions---an
essential requirement in clinical scenarios where multiple co-existing abnormalities must be precisely described.

Subsequent models like SimVLM~\citep{wang2021simvlm} and BLIP~\citep{li2023blip} introduce multi-stage pretraining and multimodal fusion mechanisms to extract both global and localized features.
However, these models are predominantly trained on natural image-caption datasets and lack integration of domain-specific knowledge, which limits their effectiveness in the medical domain where
subtle anatomical variations and specialized terminology are critical.

\textbf{2D Medical Report Generation.}
Medical image captioning demands structured, interpretable descriptions of disease patterns,  anatomical regions, and their clinical significance. Early systems  adapted CNN-RNN architectures for
chest X-rays~\citep{shin2016learning,yuan2019automatic,yin2019automatic}, but lacked control over output structure and were sensitive to data scarcity. Recent efforts have leveraged Transformers
and curriculum learning~\citep{liu2022competence} to improve generalization and mitigate reporting bias. Furthermore, memory-driven decoders incorporating relational memory have enhanced clinical
coherence in generated reports~\citep{chen2020generating}.

Several VLMs have extended to medical domains, such as BioViL    \citep{boecking2022making}, which employs attention-based fusion of visual and textual inputs. Yet, these methods often fail to
integrate structured clinical ontologies or incorporate domain priors. To bridge this gap, models like PromptMRG~\citep{jin2024promptmrg} and RGRG~\citep{tanida2023interactive_rgrg} incorporate
prompting strategies or detection-guided feature extraction to encourage sentence-level alignment. Despite these advances, most models still lack robust region-wise structural control and remain
constrained by weak domain-specific grounding.

Recent works such as MKCL~\citep{hou2023mkcl} incorporate predefined medical knowledge graphs and cross-modal contrastive learning to capture semantic relationships among findings. However, the
effectiveness of these methods in modeling spatially grounded abnormalities and aligning image features with multi-label clinical narratives remains limited.

\textbf{3D Medical Report Generation.}
Compared to 2D modalities, 3D imaging (e.g.,~CT, MRI) offers richer spatial context but poses challenges due to volumetric complexity and limited annotated data. Models such as
CT2Rep~\citep{hamamci2024ct2rep} and MedBLIP~\citep{chen2024medblip} attempt to adapt Transformers and contrastive pretraining to 3D modalities. MedBLIP, in particular, leverages 2D
vision-language pretraining and extends it to 3D via generative modeling.

Nonetheless, these methods often align representations only at the image level, overlooking the nuanced variability in radiological presentations---such as differentiating pneumonia from pulmonary
embolism---where semantic granularity and region-specific modeling are essential~\citep{wang2022medclip}. Moreover, the absence of well-defined positive-negative pairs limits the effectiveness of
standard contrastive strategies in multi-label clinical tasks, impeding accurate abnormality recognition and structured generation.

\textbf{Comparison with Existing Approaches.}
In this study, Abn-BLIP is a clinically structured vision-language framework for 3D CTPA analysis, differing from general-purpose VLMs by employing abnormality-driven visual queries aligned with a
predefined anatomical hierarchy of 32 findings. This task-specific query design, unlike the task-agnostic QFormer in BLIP~\citep{li2023blip}, enables precise region-wise feature extraction for clinical reporting. Furthermore, our abnormality-aligned contrastive learning (ACL) leverages weakly labeled text to build anomaly- and organ-specific image--text pairs, surpassing case-level
contrastive learning in capturing fine-grained, clinically relevant associations. By embedding clinical priors and enforcing structured output, Abn-BLIP improves interpretability, clinical
consistency, and semantic fidelity in automated radiology reports.

\section{Methods}

Based on clinical diagnostic guidelines for CTPA~\citep{tan2022pulmonary,bukhari2024clinical}, we identified the necessity of a systematic framework to enhance abnormality detection and structured
report generation for PE diagnosis. Accordingly, we developed a hierarchical diagnostic framework informed by the clinical expertise of radiologists from Brown University, Johns Hopkins
University, and the University of Michigan, in collaboration with emergency physicians and pulmonologists. Their combined clinical insights ensured the framework's clinical relevance, consistency,
and generalizability across diverse healthcare settings.

As illustrated in  \ref{fig::abn_bars}, the framework systematically structures the diagnostic process through a hierarchical evaluation of seven anatomical regions and 32 critical CTPA abnormalities. Within this structured approach, abnormalities are identified at a regional level and synthesized into a comprehensive diagnostic summary, facilitating precise abnormality localization and standardized.

For diagnostic model training and report generation, CTPA radiology reports were processed with a large language model (LLM)~\citep{dubey2024llama} to extract training targets. The LLM identified 32 abnormality labels ($Y$) and retrieved their corresponding text-based findings ($T$), which served as training references for both binary and textual predictions.

\begin{figure*}[!t]
\includegraphics[width=0.96\textwidth]{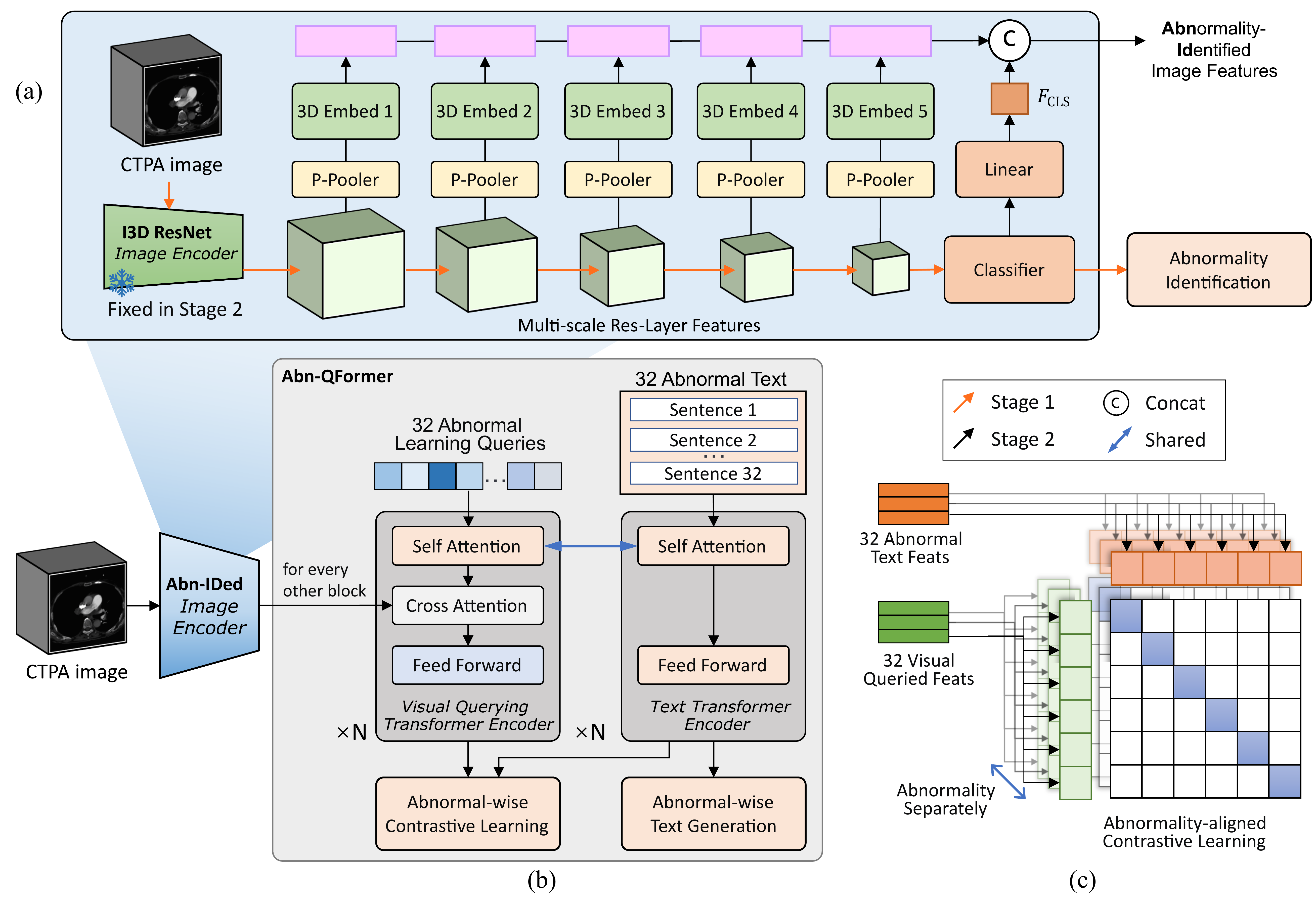}
\caption{Overview of the proposed Abn-BLIP model for CTPA abnormality diagnosis and report generation.
(a) Anatomy-guided multi-abnormality identification in Stage 1: Multi-scale abnormality-identified image feature extraction for transformer encoders.
(b) Abnormality-driven visual Querying Transformers (Abn-QFormer): Joint optimization of two objectives, enforcing abnormal queries (a set of learnable embeddings) to extract visual abnormal representations most relevant to their corresponding abnormal text descriptions.
(c) Abnormality-aligned Contrastive Learning (ACL): Achieving more fine-grained visual queried representations by aligning abnormalities.\label{fig::method_framework}}
\end{figure*}

\subsection{Anatomy-guided multi-abnormality identification}

Multi-abnormality identification in medical imaging is essential for diagnosing and monitoring the 32 abnormalities observed in CTPA scans. Unlike single-label methods that detect only one abnormality, multi-label classification enables simultaneous recognition of co-occurring conditions~\citep{ge2024dkec}, thereby improving the understanding of visual and critical relationships in cardiac and pulmonary disease. As illustrated by the orange pathway of module (a) in Fig. \ref{fig::method_framework}, Stage 1 trains an image encoder together with a multi-label classifier to detect abnormalities from CTPA images. Given an input scan $x_I$, the classifier estimates the probability $P_k$ for each abnormality class $k$.

The image encoder architecture is based on an inflated 3D (I3D) ResNet152, which extends its 2D counterpart by inflating convolutional kernels into the temporal domain~\citep{carreira2017quo}. This design preserves pretrained 2D spatial representations while capturing spatiotemporal dependencies in CTPA sequences. The model begins with a $7 \times 7 \times 3$ convolutional layer followed by max pooling, while retaining the residual connections of ResNet152 for hierarchical feature extraction. The final residual block produces a feature map of size $2048 \times 7 \times 7 \times 10$, which is passed through 3D adaptive average pooling and a 3D convolutional layer to produce logits for 32 abnormality classes. Probabilities are obtained via a sigmoid activation function, and the model is optimized using binary cross-entropy loss:
\begin{equation}
    L_{\text{cls}} = -\sum_{k=1}^{32} y_k \log P_k + (1-y_k) \log (1-P_k)
\end{equation}
where $y_k$ denotes the ground-truth label of $Y$. Optimizing $L_{\text{cls}}$ enables the model to learn robust multi-abnormality representations for comprehensive disease assessment.

To enhance visual feature representations for abnormality querying, multi-scale features ($f_I^l \in \mathbb{R}^{C_l \times H_l \times W_l}; l \in [1,5]$) are extracted from the five ResLayers of the I3D ResNet. A 3D patch-pooling module partitions each scale into non-overlapping $7 \times 7 \times 10$ sub-volumes. Average pooling is then applied within each patch, reducing spatial resolution while preserving localized spatial and semantic information. The pooled features from each scale are further embedded using 3D convolutional blocks with ReLU activation and batch normalization, compressing their dimensionality to ${C}^{\prime}_{l} \times d_v$, where $d_v$ denotes the visual feature dimension. Finally, the multi-scale embeddings are concatenated to form a unified feature representation with $M$ channels.

To integrate visual and semantic abnormality information, the predicted multi-class probabilities $P$ are projected through a linear layer into the visual feature space, forming $F_{\text{CLS}}$. This abnormality-aware embedding is then concatenated with the aggregated visual embeddings, producing a joint representation $\mathbf{v} \in \mathbb{R}^{(M+1) \times d_v}$, where $(M+1)$ denotes the total number of concatenated visual tokens.

\subsection{Abnormality-driven visual querying transformers}

We propose an Abnormality-driven visual Querying Transformers (Abn-QFormer) module to generate abnormality descriptions by aligning abnormality-specific visual features with disease-level text findings, as illustrated in module (b) of Fig \ref{fig::method_framework}.

The Abn-QFormer is composed of two complementary submodules for textual and visual information processing. The text transformer employs a Self-Attention (SA) mechanism, serving as both encoder and decoder for textual representations. The visual querying transformer extends this design by integrating Self-Attention (SA) and Cross-Attention (CA) layers, enabling interactions with visual features and producing contextual embeddings for 32 predefined abnormalities. To incorporate prior knowledge, the self-attention layers in both submodules are initialized with pre-trained BERT-base weights~\citep{devlin2018bert}, while the cross-attention layers are randomly initialized to specialize in visual learning.

For the $k$th abnormality, the textual input is represented as a tokenized sequence $ \mathbf{T}^{k} = [t_1, t_2, \dots, t_L] $, where $ L $ is the number of tokens. Each token is embedded into a fixed-dimensional vector and processed by a Multi-Head Self-Attention (MHSA) mechanism, which computes pairwise attention weights across all tokens:
\begin{equation}
 \text{SA}(\mathbf{X}) = \text{softmax}\left(\frac{\mathbf{XW}_Q \mathbf{XW}_K^\top}{\sqrt{d_k}}\right)\mathbf{XW}_V 
\end{equation}
where $ \mathbf{W}_Q $, $ \mathbf{W}_K $, and $ \mathbf{W}_V $ are learnable projection matrices, $ d_k $ is the dimensionality of the query and key vectors. The output of the attention layer is refined through a feed-forward network (FFN):
\begin{equation}
 \text{FFN}(\mathbf{X}) = \text{ReLU}(\mathbf{X} \mathbf{W}_1 + \mathbf{b}_1) \mathbf{W}_2 + \mathbf{b}_2
\end{equation}

To stabilize training, residual connections and layer normalization are applied:
\begin{equation}
\begin{split}
\mathbf{H}^l &= \text{LayerNorm}(\mathbf{H}^{l-1} + \text{MHSA}(\mathbf{H}^{l-1})) \\ 
\mathbf{H}^l &= \text{LayerNorm}(\mathbf{H}^l + \text{FFN}(\mathbf{H}^l))
\end{split}
\end{equation}

The final layer [CLS] token embedding $ \mathbf{h} $ serves as the global textual representation, capturing high-level semantic information.

The visual querying transformer extends textual processing by incorporating 32 learnable query embeddings:
\begin{equation}
 \mathbf{Q}_{\text{abn}} = [\mathbf{Q}_1, \mathbf{Q}_2, \dots, \mathbf{Q}_{32}], \mathbf{Q}_i \in \mathbb{R}^{1 \times d} 
\end{equation}

Each query is implemented as a trainable parameter that extracts abnormality-relevant features from the visual embeddings. The self-attention layers facilitate intra-query interactions, capturing contextual relationships among queries, while the cross-attention layers enable dynamic interactions with the visual features $ \mathbf{v} $, thereby aligning the queried attention with the corresponding abnormalities:
\begin{equation}
\text{CA}(\mathbf{X}, \mathbf{v}) = \text{softmax}\left(\frac{\mathbf{XW}_Q (\mathbf{vW}_K)^\top}{\sqrt{d_k}}\right)(\mathbf{vW}_V)
\end{equation}
where $ \mathbf{X} $ represents the updated query embeddings. $ \mathbf{W}_Q $, $ \mathbf{W}_K $, and $ \mathbf{W}_V $ are learnable projection matrices of CA layers. 

The Multi-Head Cross-Attention (MHCA) layers enhance the ability of queries to selectively attend to abnormality-relevant visual regions. By combining self-attention and cross-attention, the model captures both intra-query dependencies and cross-modal alignments, progressively refining the query embeddings across layers. The iterative querying process is formally defined as:
\begin{equation}
\begin{split}
\mathbf{Z}^l &= \text{LayerNorm}(\mathbf{Z}^{l-1} + \text{MHSA}(\mathbf{Z}^{l-1})) \\ 
\mathbf{Z}^l &= \text{LayerNorm}(\mathbf{Z}^l + \text{MHCA}(\mathbf{Z}^l, \mathbf{v})) \\ 
\mathbf{Z}^l &= \text{LayerNorm}(\mathbf{Z}^l + \text{FFN}(\mathbf{Z}^l))
\end{split}
\end{equation}

A shared transformer backbone across textual and visual modules improves parameter efficiency and facilitates diverse cross-modal tasks. The encoder employs bi-directional self-attention for representation learning, while the decoder utilizes causal self-attention for sequence generation. Embedding layers, cross-attention layers, and feed-forward networks (FFNs) are consistently applied in both encoding and decoding stages.

\subsection{Abnormality-aligned bootstrapping language-image pre-training}

Our abnormality-driven training framework leverages image–text pairs to optimize two complementary objectives: Abnormality-aligned Contrastive Learning (ACL) and Abnormal-Grounded Text Generation (ATG). Together, these objectives enable fine-grained cross-modal alignment and enhance the model’s ability to generate abnormality-specific descriptions.

ACL maximizes mutual information between image and text representations by aligning their embeddings in a shared feature space. This strategy strengthens cross-modal understanding by linking abnormality-specific visual queries with their corresponding textual findings. Concretely, ACL employs 32 visual query embeddings ($\mathbf{Z}$) from the transformer-based visual encoder and 32 [CLS] token embeddings ($\mathbf{h}$) from the text encoder, ensuring robust and discriminative cross-modal representation learning.

The alignment process employs a contrastive loss to jointly optimize image-to-text and text-to-image similarities, following the InfoNCE formulation~\citep{oord2018representation}. For the $k$-th abnormality, the image-to-text alignment is defined as:
\begin{equation}
\mathcal{L}^k_{\text{I2T}} = -\frac{1}{N} \sum_{i=1}^N \text{softmax}(P^k) \log \frac{\exp(\Phi(\mathbf{Z}_i^k, \mathbf{h}_i^k) / \tau)}{\sum_{j=1}^N \exp(\Phi(\mathbf{Z}_i^k, \mathbf{h}_j^k) / \tau)}
\end{equation}
where $N$ denotes the batch size, $\Phi$ represents cosine similarity, and $\tau$ is a temperature scaling factor. $P^k$ indicates the predicted abnormality probabilities, which serve as soft labels for cross-modal alignment. $\mathbf{Z}^k$ and $\mathbf{h}^k$ correspond to the normalized visual and textual embeddings of the $k$-th abnormality. The text-to-image loss is defined in an analogous manner:
\begin{equation}
\mathcal{L}^k_{\text{T2I}} = -\frac{1}{N} \sum_{i=1}^N \text{softmax}(P^k) \log \frac{\exp(\Phi(\mathbf{h}_i^k, \mathbf{Z}_i^k) / \tau)}{\sum_{j=1}^N \exp(\Phi(\mathbf{h}_i^k, \mathbf{Z}_j^k) / \tau)}
\end{equation}

The final ACL loss aggregates bidirectional contrastive losses across all 32 abnormalities, as shown in (c) of  \ref{fig::method_framework} ensuring a balanced alignment between modalities:

\begin{equation}
\mathcal{L}_{\text{ACL}} = \frac{1}{2} \sum_{k=1}^{32} (\mathcal{L}^k_{\text{I2T}} + \mathcal{L}^k_{\text{T2I}})
\end{equation}

To preserve modality-specific information and prevent feature leakage, we introduce unimodal self-attention masks, ensuring independent refinement of visual queries and text embeddings. Furthermore, freezing the image encoder during training improves efficiency while leveraging in-batch negatives for enhanced negative sampling.

Abnormality-Grounded Text Generation (ATG) trains the Abn-QFormer to generate textual descriptions conditioned on visual inputs, enabling the transformation of abnormality-related visual features into coherent textual findings. The extracted visual abnormalities, captured by learned queries, are propagated to text tokens through self-attention layers with multimodal causal masks. This structured attention design constrains information flow such that queries interact only with one another, while text tokens attend to both the queries and preceding textual context.

Text generation follows an autoregressive decoding paradigm, where a special [DEC] token is used to initialize the sequence. For the $k$th abnormality, the probability of generating the text sequence  $ \mathbf{T}^k = [t_1, t_2, \dots, t_L] $ conditioned on the queried visual embedding $ \mathbf{Z}^k $ is expressed as:
\begin{equation}
    P_{\text{ATG}}^k(\mathbf{T} \vert  \mathbf{Z}) = \prod_{i=1}^L P(t_i \vert  t_{<i}, \mathbf{Z})
\end{equation}
where $ t_i $ represents the $i$th token in the sequence, and $ t_{<i} $ denotes its preceding tokens. The model is optimized with a cross-entropy loss to maximize the likelihood of the generated text:
\begin{equation}
    \mathcal{L}_{\text{ATG}} = -\frac{1}{N} \sum_{n=1}^N \sum_{k=1}^{32} \sum_{j=1}^L \log P_{\text{ATG}}^k(t^n_j \vert  t^n_{<j}, \mathbf{Z}^n)
\end{equation}

To enhance generalization and reduce overfitting, we apply label smoothing with a factor of 0.1. By jointly optimizing ACL, ATG, and multi-label abnormality classification, our pre-training framework achieves robust cross-modal alignment while retaining the flexibility to generate detailed abnormality-specific descriptions.

\subsection{Inference for study report}

For abnormality description generation, Abn-QFormer adopts an encoder–decoder architecture with cross-modal attention to align visually queried features with tokenized textual outputs, producing 32 distinct abnormality descriptions.

To construct a comprehensive CTPA findings report, we integrate a large language model (LLM) that systematically organizes abnormality-specific descriptions across seven anatomical regions, synthesizing them into a cohesive final report. For patients with multiple scans acquired under varying imaging parameters, a LLaMA3-based report-writing agent~\citep{dubey2024llama} aggregates and summarizes abnormality-specific observations at the regional level (Supplementary A.1). To ensure clinical adherence and highlight key diagnostic insights, tailored prompt engineering is applied within a structured radiology report format.

\section{Experiments and results}

\subsection{Datasets}
To assess the effectiveness of the proposed method across multiple clinical tasks, we conducted experiments on two CTPA datasets paired with radiology reports: (1) INSPECT~\citep{huang2023inspect} from Stanford University and (2) a retrospective CTPA dataset from Brown University Health (BUH).

The INSPECT dataset~\citep{huang2023inspect}, collected at Stanford Medicine between 2000 and 2021, comprises 23,248 CTPA scans from 19,402 patients at risk for PE. It includes the impression sections of radiology reports, providing radiologist-authored diagnostic descriptions and interpretations.

The BUH dataset includes patients who underwent CTPA imaging between 2015 and 2019, with some patients having multiple follow-up scans. In total, it consists of 59,754 image–report pairs from 19,565 patients. The two datasets were combined and randomly partitioned into training, validation, and testing sets at a 7:1:2 ratio. This study was approved by the Lifespan Institutional Review Board 3 (Ref.~[1791856-20]; Project Code 214421), with informed consent waived due to the retrospective use of de-identified imaging and clinical data. All participants were over 18 years of age.

\subsubsection{Image preprocessing}
CTPA scans were preprocessed by extracting pixel data from DICOM files, standardizing spatial coordinates, and normalizing Hounsfield Units (HU). To enhance anatomical focus, lung regions were segmented and cropped with a 20 mm margin~\citep{lungregion}. Axial slices were resampled to an in-plane resolution of 1.5 mm and an out-of-plane resolution of 3 mm, then padded and cropped to $224 \times 224 \times 160$. HU values were clipped to the range [$-1000$, $1000$] and subsequently normalized to [0,1].

\subsection{Implementation details}

The training procedure is divided into two stages to effectively learn both visual and semantic representations for downstream abnormality-guided report generation. In the first stage, a multi-label abnormality classification model is trained by optimizing the parameters of a 3D image encoder based on the Inflated 3D ResNet-152 (I3D) architecture. This encoder is initialized with pre-trained weights from Merlin~\citep{blankemeier2024merlin}, a vision–language model designed for 3D CT analysis. Pre-training on a large-scale corpus of structured electronic health records and unstructured radiology reports enables the encoder to capture spatially and semantically enriched pathological patterns.

During this stage, the I3D ResNet encoder is fine-tuned to learn discriminative visual features for simultaneous identification of multiple abnormalities. Supervision is provided through binary cross-entropy loss across 32 extracted PE-related abnormality labels.

In the second stage, the image encoder is frozen to preserve the learned visual representations. Training then focuses on the remaining components of the framework, including the multi-scale image feature embeddings and the multimodal transformer layers within Abn-QFormer. This stage enhances the alignment between visual features and abnormality-specific semantic tokens, thereby enabling accurate and context-aware report generation.

For each input CTPA image, the image encoder generates multi-scale feature embeddings with abnormality probability distributions of size $257 \times 1408$. The Abn-QFormer module employs 32 learnable queries, each corresponding to a specific abnormality with a feature dimension of 768. These queries extract 32 distinct abnormality-specific visual features, each represented as a 256-dimensional vector, capturing fine-grained abnormality patterns.  

The transformer backbone consists of 12 hidden layers to support robust multimodal fusion. Training and validation are performed on two NVIDIA RTX A6000 GPUs. The model is optimized with AdamW at a learning rate of $1 \times 10^{-5}$, a batch size of 20, and a maximum of 27 epochs, ensuring stable convergence and strong performance. During inference, the pipeline proceeds in two sequential stages: abnormality detection followed by report generation. The abnormality detection stage requires approximately 1.5 GB of GPU memory, while the report generation stage consumes up to 16 GB. Despite the computational complexity, the end-to-end inference process remains efficient, with an average processing time of about one minute per case.

\subsection{Abnormality diagnosis results}

We evaluate Abn-BLIP's diagnostic performance using accuracy (ACC), area under the receiver operating characteristic curve (AUC), sensitivity (Sen.), and specificity (Spe.). Table \ref{tab::abn_detect} presents a comparative analysis of our method against state-of-the-art (SOTA) approaches for CTPA abnormality classification. As benchmarks, we select two leading medical VLMs tailored for 3D imaging: M3D~\citep{bai2024m3d} and RadFM~\citep{wu2023towards}, both of which adopt visual question answering (VQA) for abnormality detection. In these evaluations, the VLMs are prompted with 32 structured queries, each corresponding to a specific abnormality in the format:

``Is there any indication of {$<$Anomaly name$>$} in this image? (This is a true or false question, please answer `Yes' or `No')''. 

Among the compared models, M3D achieves an ACC of 0.895 but performs poorly in terms of AUC (0.499), sensitivity (0.011), and F1-score (0.479), reflecting a strong bias toward negative cases and limited ability to detect true abnormalities. RadFM exhibits the weakest overall performance, with an ACC of 0.480, AUC of 0.495, sensitivity of 0.485, specificity of 0.500, and F1-score of 0.303, suggesting insufficient discriminatory power. In contrast, our proposed method attains the highest AUC (0.773) and F1-score (0.653), with an ACC of 0.896, sensitivity of 0.384, and specificity of 0.932. Supplementary Table A.1 further details Abn-BLIP’s classification performance across all 32 abnormalities. These results highlight the model’s ability to capture fine-grained abnormality features, enabling more precise and reliable CTPA abnormality descriptions.

\begin{table}[!t]
    \caption{Comparison of current 3D medical VLMs on a combined testing set using multi-label classification metrics. The highest performances are highlighted in bold. }\label{tab::abn_detect}
    \small
    \centering
    \renewcommand\arraystretch{1.1}
    \setlength{\tabcolsep}{1.1 mm}{
    \begin{tabular}{lc|cccccc}
    \toprule[ 1.5 pt]
        \multicolumn{2}{l|}{\textbf{Methods}} & ACC & AUC & Sen. & Spe. & Precision & F1 \\ 
        \midrule [ 0.7 pt]
        \multicolumn{2}{l|}{M3D (\citeyear{bai2024m3d})} & 0.895 & 0.499 & 0.011 & \textbf{0.987} & 0.079 & 0.479 \\ 
        \multicolumn{2}{l|}{RadFM (\citeyear{wu2023towards})} & 0.480 & 0.495 &  \textbf{0.485} & 0.500 & 0.119 & 0.303 \\ 
        \multicolumn{2}{l|}{CT-CHAT (\citeyear{hamamci2024developing})} & 0.307 & 0.500 & 0.719 & 0.281 & 0.105 & 0.189  \\ 
        \multicolumn{2}{l|}{Abn-BLIP (Ours)} & \textbf{0.896} & \textbf{0.773} & 0.384 & 0.932 & \textbf{0.385} & \textbf{0.653} \\ 
        \bottomrule[ 1.5 pt]
    \end{tabular}
    }
\end{table}

\begin{table}[!t]
    \caption{Comparison of PE diagnosis performance. }\label{tab::pe_detect}
    \small
    \centering
    \renewcommand\arraystretch{1.1}
    \setlength{\tabcolsep}{1.1 mm}{
    \begin{tabular}{lc|cccccc}
    \toprule[ 1.5 pt]
        \multicolumn{2}{l|}{\textbf{Methods}} & ACC & AUC & Sen. & Spe. & Precision & F1 \\ 
        \midrule [ 0.7 pt]
        \multicolumn{2}{l|}{M3D (\citeyear{bai2024m3d})} & 0.795 & 0.500 & 0.003 & \textbf{0.997} & 0.242 & 0.446 \\ 
        \multicolumn{2}{l|}{RadFM (\citeyear{wu2023towards})} & 0.549 & 0.490 &  0.390 & 0.590 & 0.195 & 0.468 \\ 
        \multicolumn{2}{l|}{CT-CHAT (\citeyear{hamamci2024developing})} & 0.797 & 0.500 & 0.000 & 1.000 & 0.500 & 0.444  \\ 
        \multicolumn{2}{l|}{PENet (\citeyear{huang2020penet})} & 0.212 & 0.513 &  \textbf{0.984} & 0.015 & 0.203 &  0.183 \\ 
        \multicolumn{2}{l|}{Abn-BLIP (Ours)} & \textbf{0.838} & \textbf{0.732} & 0.274 & 0.982 & \textbf{0.792} & \textbf{0.656} \\ 
        \bottomrule[ 1.5 pt]
    \end{tabular}
    }
\end{table}

\begin{table}[!t]
    \caption{Diagnosis performance for the 7 anatomical regions. } \label{tab::abn_detect_region}
    \small
    \centering
    \renewcommand\arraystretch{1.1}
    \setlength{\tabcolsep}{1 mm}{
    \begin{tabular}{l|cccccc}
    \toprule[ 1.5 pt]
        \textbf{Regions} & ACC & AUC & Sen. & Spe. & Precision & F1 \\ 
        \midrule [ 0.7 pt]
        Pulmonary Arteries & 0.904 & 0.739 & 0.289 & 0.976 & 0.595 & 0.662 \\ 
        Lungs and Airways & 0.856 & 0.775 & 0.436 & 0.886 & 0.332 & 0.641 \\ 
        Pleura & 0.943 & 0.925 & 0.638 & 0.950 & 0.557 & 0.779 \\ 
        Heart & 0.909 & 0.838 & 0.488 & 0.936 & 0.419 & 0.697 \\ 
        Mediastinum and Hila & 0.867 & 0.778 & 0.434 & 0.912 & 0.363 & 0.659 \\ 
        \makecell[l]{Chest Wall and \\ \hspace{1em} Lower Neck} & 0.955 & 0.660 & 0.133 & 0.978 & 0.146 & 0.557 \\ 
        Bones & 0.957 & 0.701 & 0.170 & 0.978 & 0.170 &  0.574 \\ 
        \bottomrule[ 1.5 pt]
    \end{tabular}
    }
\end{table}

Table~\ref{tab::abn_detect_region} urther reports Abn-BLIP's diagnostic performance across the seven anatomical regions defined in our structured framework. The model demonstrated strong diagnostic capabilities in most regions, with the highest AUC observed in the Pleura (0.925), followed by the Heart (0.838) and Lungs and Airways (0.775). In terms of F1-score, the Pleura (0.779), Heart (0.697), and Pulmonary Arteries (0.662) achieved the top three values, reflecting robust detection in clinically critical regions. While the Chest Wall and Bones regions exhibited high specificity (both 0.978), their sensitivity remained relatively low (0.133 and 0.170 respectively), likely due to the lower prevalence and subtle imaging characteristics of related abnormalities. These results validate the regional adaptability of Abn-BLIP, highlighting its ability to accurately capture abnormal patterns in both high-density vascular regions and more complex thoracic structures.

For PE diagnosis comparison in Table \ref{tab::pe_detect}, M3D achieves very high specificity (0.997) but extremely low sensitivity (0.003), indicating a strong bias toward negative cases. RadFM shows a more balanced sensitivity (0.390) and specificity (0.590), but overall performance remains limited (ACC: 0.549, AUC: 0.490). PENet~\citep{huang2020penet} reaches the highest sensitivity (0.984) but produces excessive false positives (specificity: 0.015, ACC: 0.212), likely due to distributional differences in its training data that bias predictions toward high-risk cases. In contrast, Abn-BLIP delivers the most robust overall performance, achieving the highest ACC (0.838), AUC (0.732), and F1-score (0.656), along with strong specificity (0.982). Although its sensitivity (0.274) is moderate, the model effectively balances false positives and false negatives, making it a more reliable approach for PE diagnosis in CTPA analysis.

The variations in specificity and sensitivity across methods primarily arise from class imbalance and dataset distribution. For example, M3D favors the majority class (normal findings), resulting in high specificity but poor abnormality detection. Conversely, RadFM and CT-CHAT tend to over-predict abnormalities, which improves sensitivity but reduces specificity. The limited performance of medical VLMs in PE diagnosis is further constrained by the subtle imaging features of PE and the lack of task-specific optimization for CTPA. PENet’s performance is heavily influenced by domain gaps, including differences in acquisition protocols and patient populations, as well as suboptimal decision thresholds. By comparison, Abn-BLIP achieves a better balance between sensitivity and specificity, as reflected in its superior AUC, enabling more accurate and reliable abnormality detection---an essential step toward generating high-quality clinical reports (see Table \ref{tab::compare_report}).

\begin{table*}[!t]
    \caption{Natural Language Generation (NLG) metrics comparison on captioning- and learning-based report generation. }\label{tab::compare_report}
    \small
    \centering
    \renewcommand\arraystretch{1.1}
    \setlength{\tabcolsep}{3.1 pt}{
    \begin{tabular}{ll|cccccc|cccccc}
    \toprule[ 1.5 pt]
        \multicolumn{2}{l|}{\textbf{Datasets}} & \multicolumn{6}{c|}{\textbf{BUH}} & \multicolumn{6}{c}{\textbf{INSPECT}}   \\ 
        \midrule [ 0.7 pt]
        Methods & Prompt & BL-1 & BL-4 & RG-1 & RG-L & MT & BERT-F1 & BL-1 & BL-4 & RG-1 & RG-L & MT & BERT-F1 \\ 
        \midrule [ 0.7 pt]
        
        RadFM (\citeyear{wu2023towards}) & Cap & 0.178 & 0.017 & 0.159 & 0.099 & 0.148 & 0.825 & 0.136 & 0.007 & 0.105 & 0.077 & 0.091 & 0.827 \\ 
        RadFM (\citeyear{wu2023towards}) & Cap + Region & 0.208 & 0.097 & 0.270 & 0.222 & 0.411 & 0.845 & 0.375 & 0.207 & 0.358 & 0.331 & 0.481 & 0.871 \\
        RadFM (\citeyear{wu2023towards}) & Cap + Region + Oneshot & 0.209 & 0.099 & 0.261 & 0.209 & 0.374 & 0.826 & 0.389 & 0.223 & 0.399 & 0.355 & 0.471 & 0.853 \\ 
        
        M3D (\citeyear{bai2024m3d}) & Cap & 0.170 & 0.010 & 0.136 & 0.090 & 0.104 & 0.817 & 0.081 & 0.003 & 0.088 & 0.068 & 0.061 & 0.807 \\ 
        M3D (\citeyear{bai2024m3d}) & Cap + Region & 0.192 & 0.025 & 0.208 & 0.125 & 0.190 & 0.825 & 0.162 & 0.015 & 0.142 & 0.103 & 0.116 & 0.787 \\ 
        M3D (\citeyear{bai2024m3d}) & Cap + Region + Oneshot & 0.219 & 0.074 & 0.164 & 0.125 & 0.158 & 0.826 & 0.101 & 0.038 & 0.122 & 0.104 & 0.100 & 0.822 \\ 
        
        CT-CHAT (\citeyear{hamamci2024developing}) & Cap & 0.079 & 0.004 & 0.124 & 0.079 & 0.164 & 0.796 & 0.109 & 0.005 & 0.144 & 0.085 & 0.157 & 0.824 \\ 
        CT-CHAT (\citeyear{hamamci2024developing}) & Cap + Region & 0.082 & 0.003 & 0.129 & 0.080 & 0.162 & 0.797 & 0.111 & 0.004 & 0.147 & 0.087 & 0.150 & 0.825 \\ 
        CT-CHAT (\citeyear{hamamci2024developing}) & Cap + Region + Oneshot & 0.198 & 0.106 & 0.291 & 0.233 & 0.370 & 0.827 & 0.386 & 0.248 & 0.422 & 0.378 & 0.479 & 0.871   \\ 
        
        % \midrule [ 0.7 pt]
        MedBlip (\citeyear{chen2024medblip}) & Contrastive learning & 0.109 & 0.069 & 0.179 & 0.144 & 0.279 & 0.829 & 0.250 & 0.203 & 0.393 & 0.344 & 0.514 & 0.892 \\ 
        CT2Rep (\citeyear{hamamci2024ct2rep}) & Memory-driven & 0.188 & 0.003 & 0.410 & 0.384 & 0.382 & 0.821 & 0.140 & 0.003 & \textbf{0.678} & \textbf{0.677} & 0.519 & 0.862 \\ 
        Abn-BLIP (Ours) & Abnormal diagnosis &  \textbf{0.525} & \textbf{0.349} & \textbf{0.504} & \textbf{0.440} & \textbf{0.550} & \textbf{0.910} & \textbf{0.652} & \textbf{0.532} & 0.630 & 0.588 & \textbf{0.704} & \textbf{0.937} \\ 
        \bottomrule[ 1.5 pt]
    \end{tabular}
    }
\end{table*}

\begin{table*}[!t]
    \caption{Clinical Efficacy (CE) metrics comparison between baseline models and the proposed Abn-BLIP model.}\label{tab::compare_report_ce}
    \small
    \centering
    \renewcommand\arraystretch{1.1}
    \setlength{\tabcolsep}{11 pt}{
    \begin{tabular}{ll|ccc|ccc}
    \toprule[ 1.5 pt]
        \multicolumn{2}{l|}{\textbf{Datasets}} & \multicolumn{3}{c|}{\textbf{BUH}} & \multicolumn{3}{c}{\textbf{INSPECT}}   \\ 
        \midrule [ 0.7 pt]
        Methods & Prompt & Precision & Recall\ & F1 & Precision & Recall\ & F1  \\ 
        \midrule [ 0.7 pt]
                
        RadFM (\citeyear{wu2023towards}) & Cap & 0.239 & 0.297 & 0.221 & 0.168 & 0.176 & 0.128 \\ 
        RadFM (\citeyear{wu2023towards}) & Cap + Region & 0.178 & 0.096 & 0.096 & 0.133 & 0.085 & 0.075  \\
        RadFM (\citeyear{wu2023towards}) & Cap + Region + Oneshot & 0.174 & 0.392 & 0.216 & 0.118 & 0.240 & 0.140 \\ 
        
        M3D (\citeyear{bai2024m3d}) & Cap & 0.221 & 0.169 & 0.157 & 0.161 & 0.119 & 0.108 \\ 
        M3D (\citeyear{bai2024m3d}) & Cap + Region & 0.187 & 0.197 & 0.155 & 0.147 & 0.136 & 0.114 \\ 
        M3D (\citeyear{bai2024m3d}) & Cap + Region + Oneshot & 0.161 & 0.256 & 0.164 & 0.152 & 0.201 & 0.147 \\ 

        CT-CHAT (\citeyear{hamamci2024developing}) & Cap & 0.208 & \textbf{0.574} & 0.278 & 0.165 & 0.321 & 0.194 \\ 
        CT-CHAT (\citeyear{hamamci2024developing}) & Cap + Region & 0.186 & 0.461 & 0.238 & 0.170 & 0.293 & 0.188  \\ 
        CT-CHAT (\citeyear{hamamci2024developing}) & Cap + Region + Oneshot & 0.199 & 0.559 & 0.271 & 0.161 & 0.478 & 0.221  \\ 
        % \midrule [ 0.7 pt]
        MedBlip (\citeyear{chen2024medblip}) & Contrastive learning & 0.110 & 0.078 & 0.070 & 0.097 & 0.067 & 0.065 \\ 
        CT2Rep (\citeyear{hamamci2024ct2rep}) & Memory-driven & 0.106 & 0.114 & 0.105 & 0.170 & 0.193 & 0.169 \\ 
        Abn-BLIP (Ours) & Abnormal diagnosis & \textbf{0.425} & 0.552 & \textbf{0.429} & \textbf{0.423} & \textbf{0.482} & \textbf{0.398} \\ 
        \bottomrule[ 1.5 pt]
    \end{tabular}
    }
\end{table*}

\subsection{CTPA report generation results}

We compare Abn-BLIP against SOTA medical VLMs and 3D medical report generation frameworks. Given the sensitivity of report generation to prompt design, we evaluate multiple prompting strategies for VLMs (see Supplementary A.5), including general captioning, organ-specific lists, and one-shot examples designed to elicit targeted abnormality descriptions. Additionally, we compare with representative 3D medical report generation models, CT2Rep~\citep{hamamci2024ct2rep} and MedBlip~\citep{chen2024medblip}, which leverage contrastive learning and memory-driven frameworks for cross-domain image-to-report generation.

Report quality is assessed using standard Natural Language Generation (NLG) metrics, including  BLEU (Bilingual
Evaluation Understudy) to evaluate fluency and adequacy based on n-gram overlap~\citep{lin-och-2004-orange}, ROUGE (Recall-Oriented Understudy for Gisting Evaluation) to assess content overlap for
summarization~\citep{lin-2004-rouge}, METEOR (Metric for Evaluation of Translation with Explicit ORdering) to incorporate unigram matching, semantic similarity, and morpheme
analysis~\citep{banarjee2005}, and BERTScore, which leverages pre-trained language model embeddings to measure semantic similarity~\citep{BERT-score}.

Table \ref{tab::compare_report} presents model performance on the BUH and INSPECT testing sets, evaluated using BLEU (BL-1, BL-4), ROUGE (RG-1, RG-L), METEOR (MT), and BERT F1-score. Across both datasets, Abn-BLIP outperforms all baselines, demonstrating its superior capacity to generate clinically relevant and well-structured reports. On the BUH dataset, Abn-BLIP achieves a BLEU-1 score of 0.525, BLEU-4 of 0.349, ROUGE-1 of 0.504, ROUGE-L of 0.440, and a BERT F1-score of 0.910. On the INSPECT dataset, it attains a BLEU-1 of 0.652, BLEU-4 of 0.532, ROUGE-1 of 0.630, ROUGE-L of 0.588, and a BERT F1-score of 0.937, confirming robustness across diverse cohorts.

VLM-based models show variable performance depending on prompting strategies. Both RadFM and M3D improve with regional prompts and one-shot learning, with RadFM consistently outperforming M3D across most NLG metrics. For example, RadFM achieves stronger BLEU-4 scores (0.099 vs. 0.074 on BUH, 0.223 vs. 0.038 on INSPECT) and higher ROUGE values, reflecting its better capacity to generate detailed disease descriptions. However, despite such improvements, all VLM-based methods (RadFM, M3D, and CT-CHAT) remain substantially weaker than Abn-BLIP. This performance gap underscores the challenges VLMs face in producing structured, clinically coherent radiology reports—particularly in maintaining both diagnostic accuracy and linguistic quality across evaluation metrics. These findings suggest that while prompt engineering can enhance VLM output, fundamental limitations persist in their ability to meet clinical reporting standards compared to task-specific medical systems.

Supervised learning-based models, MedBlip and CT2Rep, demonstrate stronger performance than VLMs in most cases. Notably, CT2Rep achieves the highest ROUGE-1 (0.678) and ROUGE-L (0.677) on the INSPECT dataset, indicating strong summarization capabilities for key findings. Nevertheless, both models face difficulties in generating long-form, well-structured reports, limiting their overall utility. By contrast, Abn-BLIP effectively synthesizes detailed, structured findings while preserving clinical relevance and linguistic coherence, achieving state-of-the-art performance in automated radiology report generation.

The Clinical Efficacy (CE) metrics~\citep{liu2019clinically} provide a rigorous assessment of diagnostic accuracy by quantifying the alignment between generated reports and ground-truth clinical observations. As shown in Table \ref{tab::compare_report_ce}, CT-CHAT achieves the highest recall (0.574 on BUH, 0.478 on INSPECT), reflecting superior sensitivity in abnormality detection, but this comes at the expense of precision (0.186–0.208), indicating a tendency toward false-positive identifications. In contrast, the proposed Abn-BLIP model delivers superior overall performance, attaining the highest precision (0.425 on BUH, 0.423 on INSPECT) and F1 scores (0.429 on BUH, 0.398 on INSPECT), thereby achieving an optimal balance between sensitivity and specificity.

The VLM-based models exhibit heterogeneous behaviors. RadFM improves recall with one-shot prompting (0.392 on BUH, 0.240 on INSPECT) but suffers from diminished precision, while M3D maintains more consistent yet generally lower performance across all CE metrics. CT-CHAT’s pattern of high recall but low precision and F1 indicates a detection strategy that emphasizes exhaustive abnormality identification at the cost of diagnostic specificity.

MedBlip and CT2Rep show weaker performance on CE metrics, suggesting that their contrastive learning and memory-driven architectures may be less suited for clinical diagnostic tasks. Collectively, these results demonstrate that while existing VLMs achieve reasonable abnormality detection, Abn-BLIP establishes a new benchmark for clinical report generation by simultaneously optimizing sensitivity and precision, as evidenced by its superior F1 performance across both datasets.

\subsection{Image-text correlation of abnormalities}

\begin{figure}[t]
\centering
\includegraphics[width=0.5\textwidth]{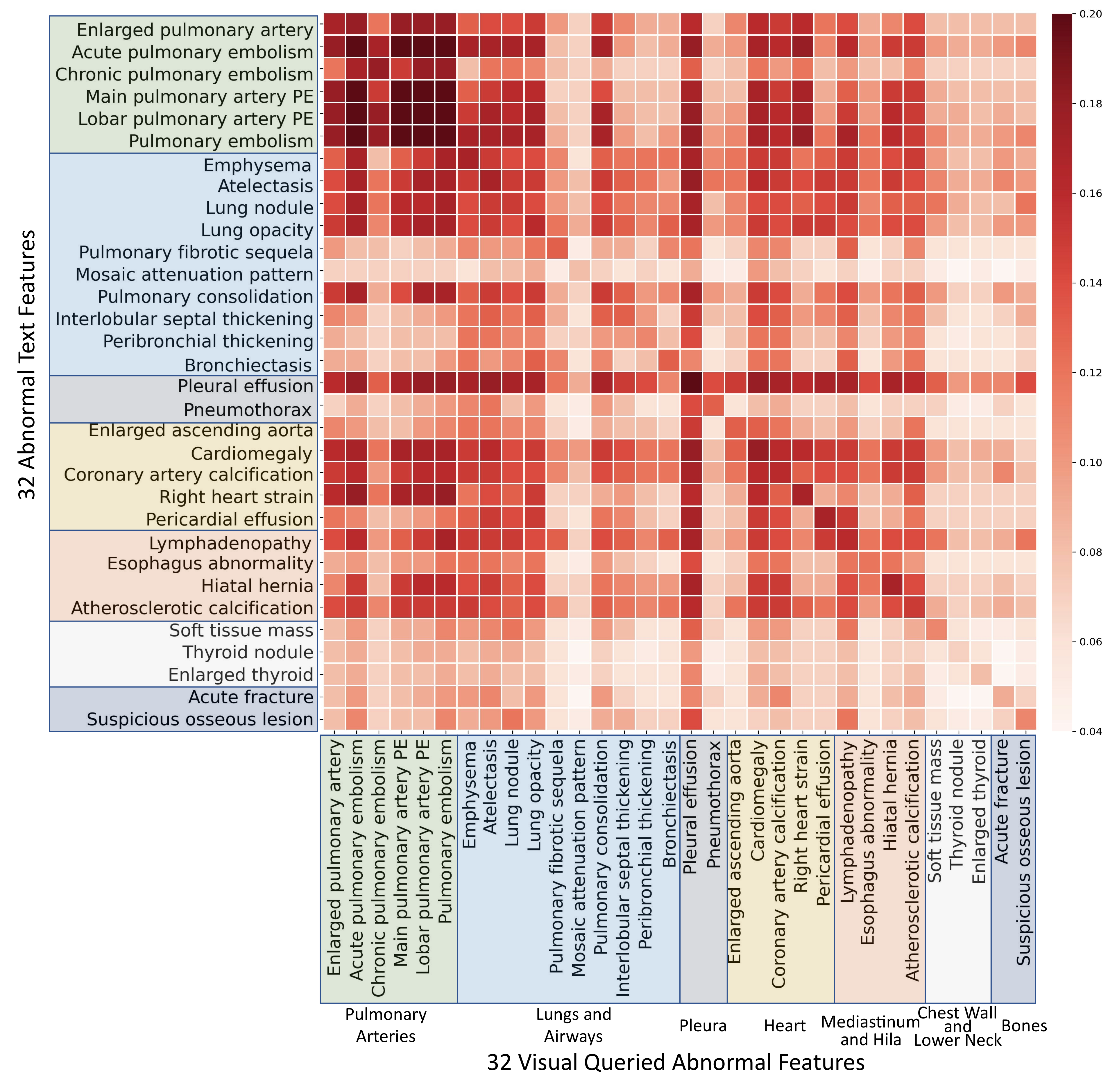}
\caption{Visualization of cross-modal cosine similarity heatmap between textual and visual features of 32 distinct CTPA abnormalities. The textual features are derived from the text descriptions of each abnormality, while the visual features are the queried representations on the corresponding images. Each cell in the heatmap indicates the similarity score between a specific abnormality's textual and visual representation, providing insights into the alignment between the two modalities}
\label{corr}
\end{figure}

Fig. \ref{corr} presents a heatmap of cross-modal cosine similarity between textual and visual features for 32 distinct CTPA abnormalities, offering insights into their anatomical relationships. High similarity scores are concentrated in the Pulmonary Arteries, Heart, and specific Lung and Airway regions, reflecting their vascular interdependence. Pulmonary arteries mediate blood flow between the heart and lungs, underscoring the clinical importance of PE-related abnormalities.

Localized high-intensity diagonal clusters within the Pulmonary Arteries region indicate strong intra-regional alignment. Abnormalities such as ``acute pulmonary embolism'' and ``pulmonary embolism'' (including ``main pulmonary artery PE'' and ``lobar pulmonary artery PE'') exhibit high similarity, consistent with their shared vascular etiology. By contrast, ``chronic pulmonary artery embolism'' shows lower similarity to acute PE variants, highlighting distinct pathophysiological processes. Acute PE typically presents with sudden vascular obstruction and hemodynamic instability, whereas chronic PE develops progressively with more subtle radiographic manifestations.

Moderate similarity is observed for ``Pleural effusion'', with off-diagonal patterns suggesting overlap with Lungs and Airways due to anatomical proximity. This finding reflects shared radiographic characteristics, as pleural effusions often co-occur with pulmonary abnormalities, reinforcing the need for contextual inter-regional analysis in diagnostic frameworks.

In contrast, abnormalities in the Chest Wall, Lower Neck, and Bones demonstrate consistently low similarity scores. Conditions such as ``suspicious osseous lesion'' and ``thyroid nodule'' show weak associations with vascular and pulmonary abnormalities, reflecting distinct diagnostic contexts and their lower prevalence in PE-related assessments. The overall correlation highlights the central role of vascular structures in PE diagnosis and emphasize the utility of cross-modal feature alignment in capturing both inter- and intra-regional relationships during CTPA analysis.

\begin{figure}[!t]
\centering
\includegraphics[width=0.5\textwidth]{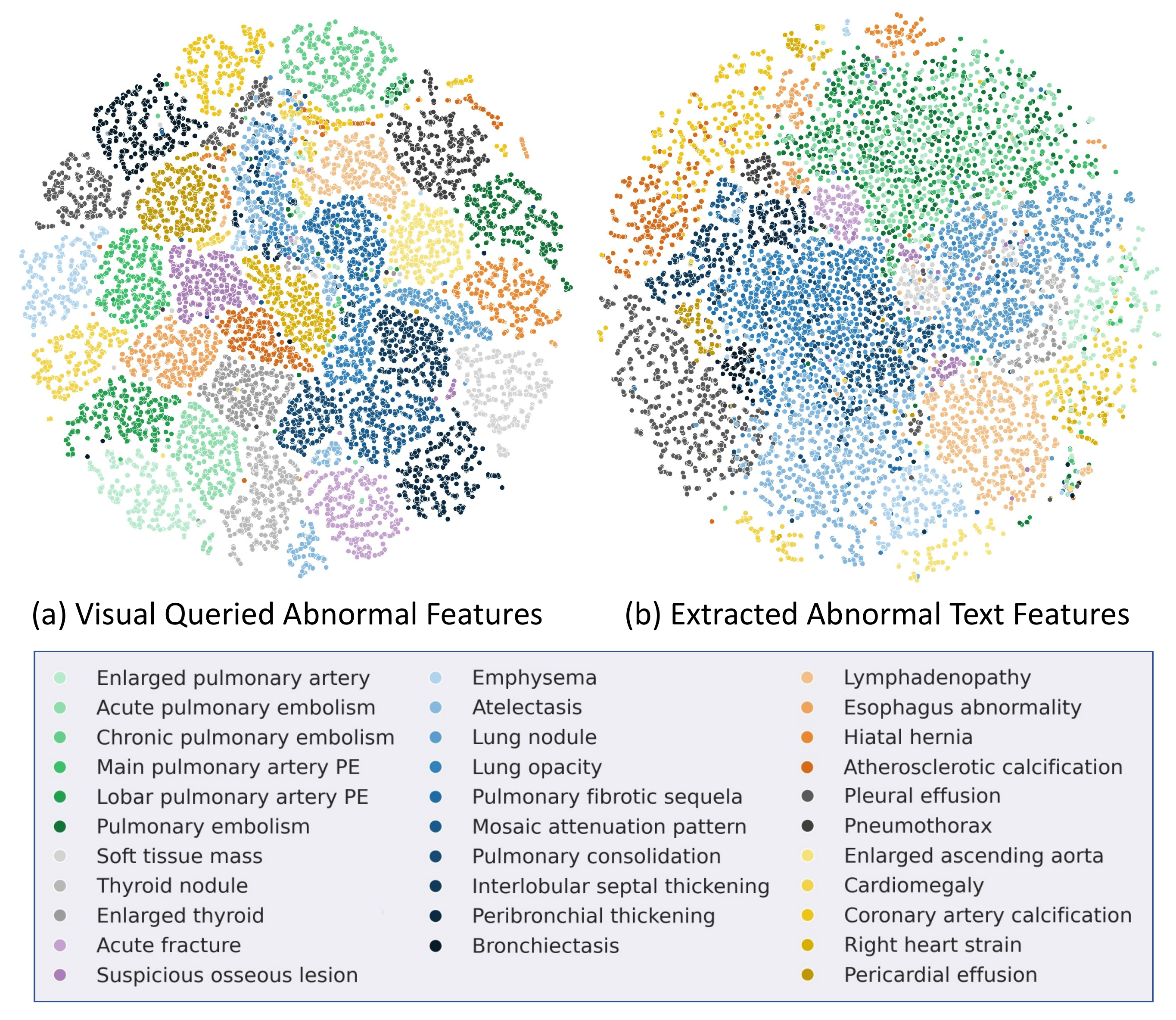}
\caption{t-SNE visualization of normalized image and text features for abnormalities. Each colored point represents one of 32 detected abnormalities, from 20,000 randomly sampled features. (a) The abnormal image features were extracted using visual querying, guided by learned abnormality-wise queries from the visual querying transformer encoder. (b) The abnormal text features were encoded by a text transformer encoder based on descriptive sentences of the abnormalities. }
\label{fig::tsne}
\end{figure}

\begin{figure*}[!ht]
\centering
\includegraphics[width=0.85\textwidth]{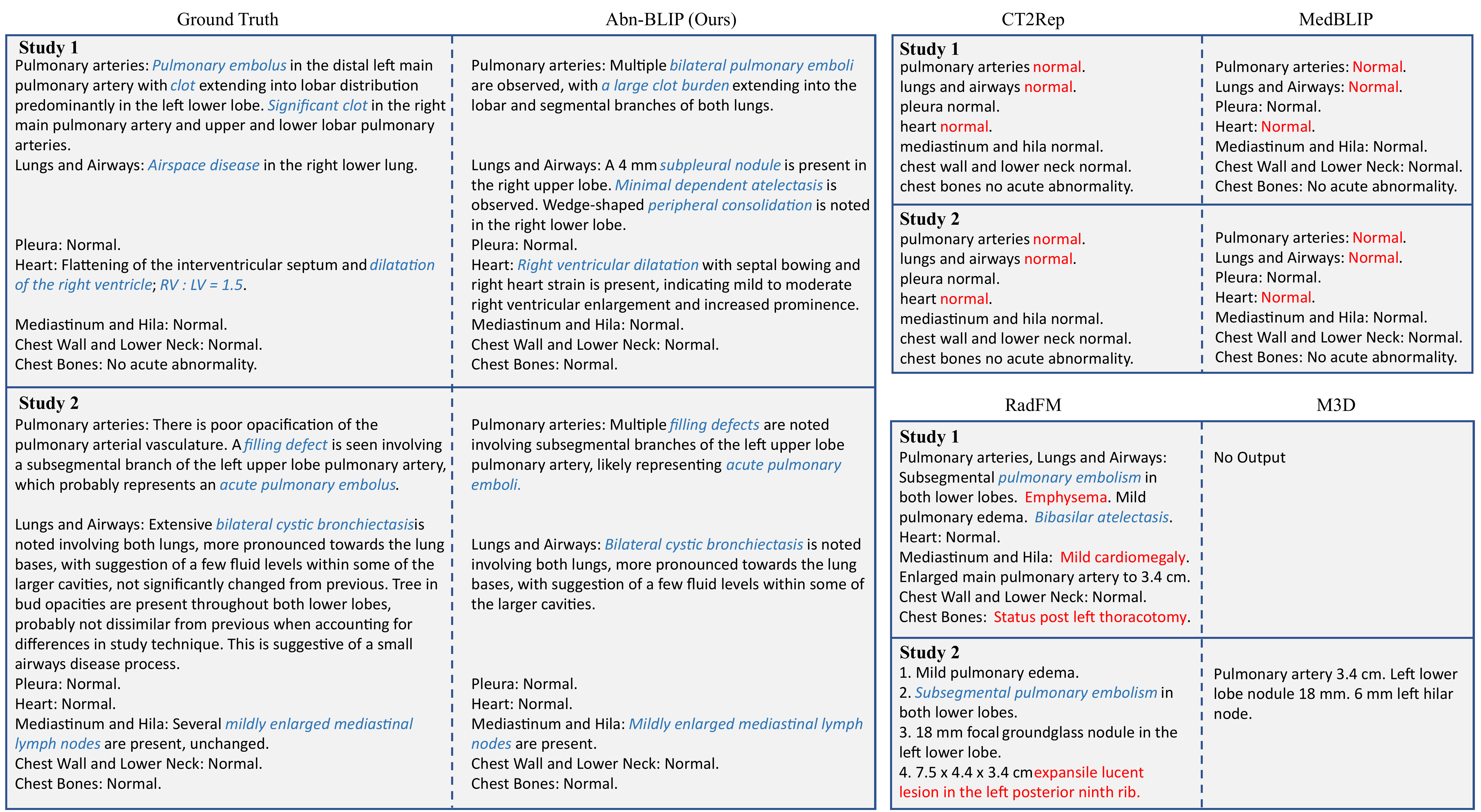}
\caption{Examples of the generated reports. Our results are compared with the ground truth, the 3D report generation methods and medical VLM methods. The blue italic text is the correct predictions corresponding to the actual reports, and the red areas indicate the untrue information in the predictions.}
\label{fig::report_sample}
\end{figure*}

\subsection{Visualization of abnormal representation}

Fig. \ref{fig::tsne} presents a t-SNE visualization of learned representations for visual (a) and textual (b) features across 32 distinct CTPA abnormalities, illustrating their clustering patterns and separability. Each point corresponds to a feature vector, with colors denoting abnormality categories.

In the visual feature plot (a), distinct clusters emerge among related abnormalities. For instance, PE-associated abnormalities (``Enlarged pulmonary artery'', ``Acute pulmonary embolism'', and ``Chronic pulmonary embolism'') cluster closely, as do lung parenchymal abnormalities (``Lung nodule'', ``Lung opacity'', and ``Pulmonary consolidation''). This suggests that visual features effectively capture morphological and textural patterns specific to each category.

In the textual feature plot (b), tighter clustering is observed at the organ level, with clear groupings in the Lungs and Airways (blue) and Pulmonary Arteries (green), indicating the textual similarities within the shared organ. Abnormalities such as ``Enlarged pulmonary artery'', ``Lymphadenopathy'', ``Esophagus abnormality'', and ``Atherosclerotic calcification'', remain well separated, reflecting their distinct semantic characteristics.

Comparing the two plots, the visual features exhibit greater separability, indicating that learnable queries enhance feature discrimination. Nevertheless, both modalities demonstrate strong discriminative capacity, underscoring the importance of integrating visual and textual representations for comprehensive abnormality characterization.

\begin{table}[t]
    \caption{Ablation studies for abnormality multi-label classification.} \label{tab::abn_detect_ablation}
    \small
    \centering
    \renewcommand\arraystretch{1.1}
    \setlength{\tabcolsep}{2 mm}{
    \begin{tabular}{lc|ccccc}
    \toprule[ 1.5 pt]
        Img-Feat & $F_{\text{CLS}}$ & ACC & AUC & Sen. & Spe. & F1 \\ 
        \midrule [ 0.7 pt]
        L5 only & - & 0.894 & 0.772 & \textbf{0.393} & 0.930 & 0.652 \\ 
        L5 only & \checkmark & 0.895 & 0.772 & 0.387 & 0.931 & 0.653 \\ 
        Multi-scale & - &  0.896 & 0.773 & 0.387 & 0.932 & \textbf{0.654}  \\ 
        Multi-scale & \checkmark & \textbf{0.896}  & \textbf{0.773}  & 0.384  & \textbf{0.932}  & 0.653 \\ 
        \bottomrule[ 1.5 pt]
    \end{tabular}
    }
\end{table}

\begin{table}[t]
\caption{Ablation studies for report generation.} \label{tab::ablation_report}
\centering
    \small
    \renewcommand\arraystretch{1.1}
    \setlength{\tabcolsep}{1.1 mm}{
        \begin{tabular}{lc|cccccc}
        \toprule[ 1.5 pt]

        \multicolumn{2}{c|}{\textbf{Ablations}}& \multicolumn{6}{c}{\textbf{Abnormality level description}}   \\ 
        % \midrule [ 0.7 pt]
        Img-Feat & $F_{\text{CLS}}$ & BL-1 & BL-4 & RG-1 & RG-L & MT & BERT-F1 \\ 
        \midrule [ 0.7 pt]
        L5 only & - & 0.414 & 0.050 & 0.649 & 0.647 & 0.632 & 0.952 \\
        L5 only & \checkmark & 0.623 & 0.096 & 0.828 & 0.826 & 0.788 & 0.977 \\
        Multi-scale & - & 0.672 & 0.107 & 0.866 & 0.865 & 0.825 & 0.982 \\ 
        Multi-scale & \checkmark & \textbf{0.677} & \textbf{0.112} & \textbf{0.874} & \textbf{0.873} & \textbf{0.831} & \textbf{0.983} \\ 
        \midrule[ 1.5 pt]
        \multicolumn{2}{c|}{\textbf{Ablations}} & \multicolumn{6}{c}{\textbf{Study level findings}}   \\ 
        Img-Feat & $F_{\text{CLS}}$ & BL-1 & BL-4 & RG-1 & RG-L & MT & BERT-F1 \\ 
        \midrule [ 0.7 pt]
        L5 only & - & 0.424 & 0.292 & 0.491 & 0.428 & 0.614 & 0.907 \\
        L5 only & \checkmark & 0.577 & 0.430 & 0.574 & 0.522 & 0.639 & 0.925 \\
        Multi-scale & - & 0.581 & 0.431 & 0.571 & 0.519 & 0.639 & 0.925 \\ 
        Multi-scale & \checkmark & \textbf{0.594} & \textbf{0.446} & \textbf{0.578} & \textbf{0.527} & \textbf{0.641} & \textbf{0.926} \\ 
                 
        \bottomrule[ 1.5 pt]
        \end{tabular}
        }
\end{table}

\subsection{Ablation study}
To evaluate the effectiveness of the proposed architectural components, we conducted ablation studies focusing on multi-scale feature fusion and the abnormal prediction embedding ($F_{\text{CLS}}$).

Table \ref{tab::abn_detect_ablation} presents multi-label abnormality classification results. Using only the fifth residual layer (L5) features without $F_{\text{CLS}}$ establishes a strong baseline, achieving an accuracy of 0.894, AUC of 0.772, sensitivity of 0.393, specificity of 0.930, and F1-score of 0.652. Adding $F_{\text{CLS}}$ slightly improves the F1-score (0.653) with minimal changes to other metrics, suggesting a modest but positive contribution. Multi-scale feature fusion alone yields more consistent improvements across all metrics (e.g., F1-score: 0.654), demonstrating its effectiveness in enhancing feature representation. However, combining both does not produce further gains, with sensitivity decreasing (0.384) and F1-score remaining at 0.653, indicating that $F_{\text{CLS}}$ provides limited additional benefit when multi-scale features are incorporated.

For report generation, Table \ref{tab::ablation_report} evaluates both abnormality- and study-level descriptions. Using L5 features alone results in weaker performance, with BLEU-1 of 0.414, BLEU-4 of 0.050, ROUGE-L of 0.647, METEOR of 0.632, and BERT-F1 of 0.952. Incorporating $F_{\text{CLS}}$ significantly improves all metrics (e.g., BLEU-1: 0.623, ROUGE-L: 0.828), reflecting its role in producing more descriptive and semantically relevant reports. Multi-scale features further enhance generation quality (e.g., BLEU-1: 0.672, ROUGE-L: 0.874), and the combination achieves the best overall results, with BLEU-1 at 0.677, BLEU-4 at 0.112, METEOR at 0.831, and BERT-F1 peaking at 0.983.

A similar trend is observed at the study level. L5 features alone yield weaker outcomes (BLEU-1: 0.424, BLEU-4: 0.092, ROUGE-L: 0.491). Incorporating $F_{\text{CLS}}$ improves contextual understanding (BLEU-1: 0.523, ROUGE-L: 0.684), while multi-scale features further boost performance (BLEU-1: 0.590, BLEU-4: 0.446). The best performance is achieved by combining both, with BLEU-1 at 0.594 and ROUGE-L at 0.872.

These findings underscore the benefits of multi-scale feature fusion in both classification and report generation. While $F_{\text{CLS}}$ contributes little to classification when multi-scale features are present, it plays a crucial role in enhancing report descriptiveness and coherence. This highlights the importance of hierarchical feature integration and contextual embedding in advancing medical image analysis.

\subsection{Qualitative analysis}
Fig. \ref{fig::report_sample} presents a comparative case study of radiology reports generated by the proposed Abn-BLIP model, existing 3D report generation methods (CT2Rep, MedBLIP), and medical VLMs (RadFM, M3D), evaluated against the ground truth. Two representative cases are shown, each illustrating distinct anatomical regions and pathological findings.

The reports of Abn-BLIP closely align with the ground truth, accurately identifying key pulmonary and cardiac abnormalities. In Study 1, it identifies bilateral pulmonary emboli with a large clot burden, subpleural nodules, peripheral consolidation, and right ventricular dilatation. In Study 2, it detects multiple filling defects consistent with acute pulmonary embolism, bilateral cystic bronchiectasis with fluid levels, and mildly enlarged mediastinal lymph nodes. 

By contrast, 3D report generation methods demonstrate substantial limitations. CT2Rep and MedBLIP frequently underreport findings, often misclassifying abnormalities as normal across multiple organ systems. Their low sensitivity, particularly for critical conditions such as pulmonary embolism, undermines clinical reliability.

The performance of medical VLMs is more variable. RadFM correctly identifies pulmonary embolism, mild pulmonary edema, and cardiomegaly but fails to detect bronchiectasis and introduces a potentially incorrect finding (status post left thoracotomy). M3D exhibits limited robustness, failing to generate output in Study 1. Although it identifies pulmonary artery enlargement and lymphadenopathy in Study 2, it misses acute pulmonary embolism and cystic bronchiectasis, resulting in incomplete diagnostic coverage.

\subsection{Expert evaluation for generated report}

To quantitatively assess the clinical quality of radiology reports generated by Abn-BLIP, we conducted a blinded evaluation involving three expert review groups, each led by board-certified radiologists and following standardized assessment criteria. A total of 60 patient cases were randomly selected from the BUH test set. For each case, reviewers were presented with both the ground-truth report and the corresponding findings generated by Abn-BLIP. Clinical quality is rated on a 5-point Likert scale (1 = lowest, 5 = highest), considering accuracy, clarity, and relevance to the reference report. In addition, reviewers assigned a 5-point confidence score reflecting their certainty in the evaluation. The assessments were performed using a previously published web-based platform for radiology report evaluation~\citep{Ma2025unified}.

\begin{figure}[!t]
\centering
\includegraphics[width=0.45\textwidth]{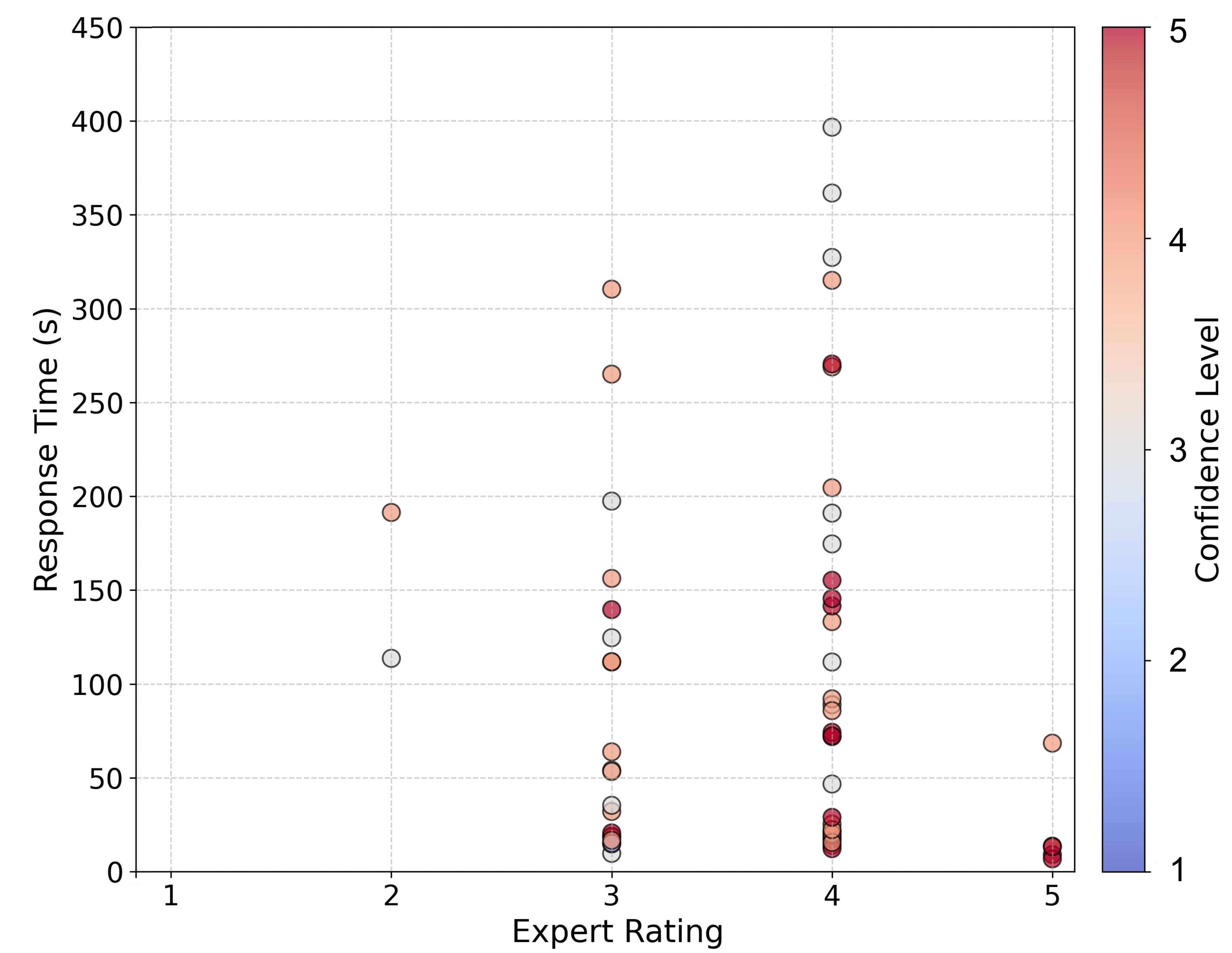}
\caption{Scatter plot of expert ratings, response times, and confidence levels in evaluating generated radiology reports.
Each dot represents a single expert assessment. The $x$-axis indicates the expert rating, the $y$-axis denotes the response time (in seconds), and color encodes the reviewer's confidence level.}
\label{fig::expert_eval}
\end{figure}

As illustrated in Fig. \ref{fig::expert_eval}, most expert ratings cluster at levels 4 and 5, indicating consistent recognition of the generated reports’ clinical accuracy and relevance. High ratings ($\geq$4) are generally associated with shorter review times (typically <150 s), suggesting that higher-quality outputs require less cognitive effort to validate. Conversely, low ratings (1–2) occur infrequently and are often accompanied by longer review times, reflecting the additional scrutiny required for reports judged less reliable.

The scatter plot’s color gradient represents reviewer confidence, ranging from 1 (``Very Low Confidence'', blue) to 5 (``Very High Confidence'', red). High-confidence assessments are predominantly aligned with high ratings, forming dense clusters in the upper-left quadrant. In contrast, lower-confidence ratings are more dispersed across mid- and low-score ranges, indicating a strong association between expert-perceived quality, confidence, and evaluation efficiency.

\begin{figure*}[!t]
\centering
\includegraphics[width=0.9\textwidth]{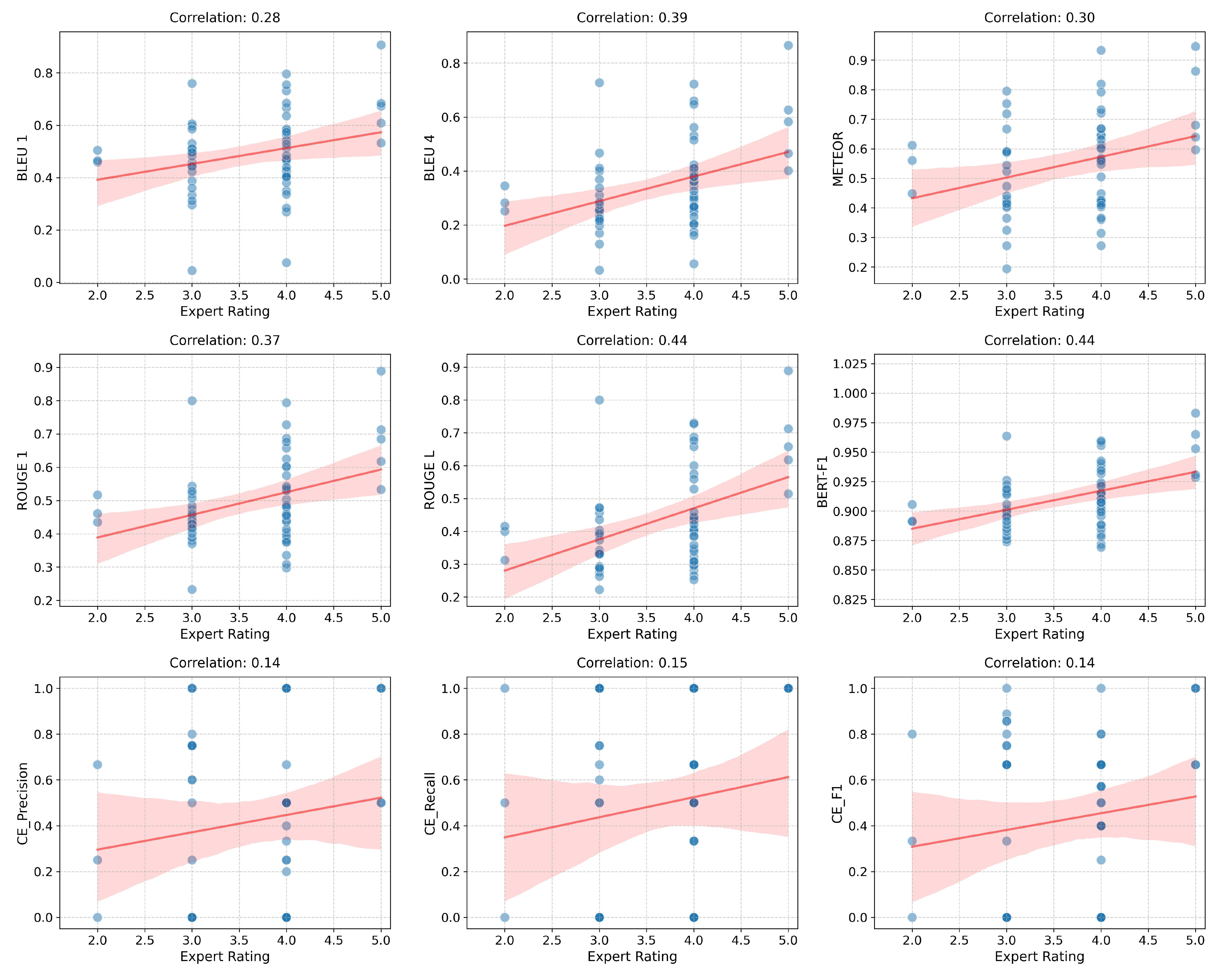}
\caption{Correlation between expert ratings and automated evaluation metrics for generated radiology reports.
Each subplot shows the relationship between expert scores and a specific evaluation metric (NLG or CE). Pearson correlation coefficients are reported, and red lines indicate linear regression fits with 95\% confidence intervals.}
\label{fig::expert_corr}
\end{figure*}

To further examine the alignment between expert assessments and automated evaluation, we analyzed the correlations between expert ratings and both NLG and CE metrics, as shown in Fig. \ref{fig::expert_corr}. Moderate positive correlations are observed with NLG metrics, particularly ROUGE-L (r $=$ 0.44) and BERTScore F1 (r $=$ 0.44), followed by ROUGE-1 (r $=$ 0.37), BLEU-4 (r $=$ 0.39), METEOR (r $=$ 0.30), and BLEU-1 (r $=$ 0.28). These results indicate that expert judgments align moderately with automated measures of textual similarity and semantic overlap, particularly those emphasizing phrase-level coherence and semantic fidelity.

The CE metrics, including CE-Precision (r $=$ 0.14), CE-Recall (r $=$ 0.15), and CE-F1 (r $=$ 0.14), exhibit only weak correlations with expert ratings. This suggests that while CE metrics provide a structured evaluation of abnormality coverage, they insufficiently capture clinically salient attributes such as narrative coherence, contextual appropriateness, and diagnostic reasoning, factors more intuitively assessed by human experts. These findings underscore the importance of expert evaluation as a critical complement to automated metrics in assessing the clinical utility of AI-generated radiology reports.

\subsection{Expert evaluation for abnormality extraction}

To evaluate the reliability of LLM-based abnormality annotation, we assessed the performance of LLaMA3~\citep{dubey2024llama} in extracting binary abnormality labels from free-text radiology reports. A total of 100 CTPA cases were randomly sampled, and LLM-extracted labels for 23 predefined PE-related abnormalities were compared against expert-annotated gold-standard labels obtained through manual review by the expert groups.

As summarized in Table \ref{tab::abn_extract}, the LLM demonstrates strong overall performance, achieving an average accuracy of 0.924, AUC of 0.879, sensitivity of 0.816, specificity of 0.942, and precision of 0.887 across all abnormalities. Perfect classification (accuracy, sensitivity, specificity, and precision $=$ 1.0) is observed for several high-prevalence and well-defined abnormalities, including Mosaic attenuation pattern, Bronchiectasis, Pleural effusion, Hiatal hernia, and Pericardial effusion.

Performance is comparatively lower for abnormalities with variable or context-dependent phrasing. For example, Chronic pulmonary embolism (Precision: 0.333, Sensitivity: 0.500) and Pulmonary consolidation (Sensitivity: 0.200) show reduced detection, highlighting challenges in disambiguating subtle or overlapping textual expressions. Nevertheless, for clinically critical findings such as Acute pulmonary embolism, Main pulmonary artery PE, and Lobar pulmonary artery PE, the LLM achieved robust sensitivity (0.710--0.958) and precision (0.590--1.000), confirming its effectiveness in capturing PE-related abnormalities.

These results indicate that the SOTA general-purpose LLMs can serve as reliable tools for automated abnormality label extraction in structured report generation pipelines, particularly for high-frequency and clearly articulated findings. However, improvements in prompt engineering and context-aware parsing are likely needed to address more nuanced or infrequently reported abnormalities.

\begin{table}[!t]
\caption{Expert Assessment of LLMs on Abnormality Extraction} \label{tab::abn_extract}
\centering
    \small
    \renewcommand\arraystretch{1}
    \setlength{\tabcolsep}{1 mm}{
        \begin{tabular}{l|ccccc}
        \toprule[ 1.5 pt]
        Abnormality & ACC & AUC & Sen. & Spe. & Precision  \\ 
        \midrule [ 0.7 pt]
        Acute pulmonary embolism & 0.730 & 0.855 & 0.710 & 1.000 & 1.000 \\ 
        Chronic pulmonary embolism & 0.940 & 0.729 & 0.500 & 0.958 & 0.333 \\ 
        Main pulmonary artery PE & 0.830 & 0.874 & 0.958 & 0.789 & 0.590 \\ 
        Lobar pulmonary artery PE & 0.760 & 0.781 & 0.930 & 0.632 & 0.656 \\ 
        Pulmonary embolism & 0.930 & 0.719 & 0.939 & 0.500 & 0.989 \\ 
        Emphysema & 0.990 & 0.950 & 0.900 & 1.000 & 1.000 \\ 
        Atelectasis & 0.940 & 0.921 & 0.842 & 1.000 & 1.000 \\ 
        Lung nodule & 0.900 & 0.849 & 0.750 & 0.947 & 0.818 \\ 
        Lung opacity & 0.710 & 0.655 & 0.310 & 1.000 & 1.000 \\ 
        Pulmonary fibrotic sequela & 0.990 & 0.833 & 0.667 & 1.000 & 1.000 \\ 
        Mosaic attenuation pattern & 1.000 & 1.000 & 1.000 & 1.000 & 1.000 \\ 
        Pulmonary consolidation & 0.800 & 0.600 & 0.200 & 1.000 & 1.000 \\ 
        Interlobular septal thickening & 0.980 & 0.895 & 0.800 & 0.989 & 0.800 \\ 
        Peribronchial thickening & 0.990 & 0.995 & 1.000 & 0.989 & 0.875 \\ 
        Bronchiectasis & 1.000 & 1.000 & 1.000 & 1.000 & 1.000 \\ 
        Pleural effusion & 1.000 & 1.000 & 1.000 & 1.000 & 1.000 \\ 
        Cardiomegaly & 0.950 & 0.968 & 1.000 & 0.936 & 0.815 \\ 
        Coronary artery calcification & 0.990 & 0.975 & 0.950 & 1.000 & 1.000 \\ 
        Right heart strain & 0.920 & 0.800 & 0.600 & 1.000 & 1.000 \\ 
        Pericardial effusion & 1.000 & 1.000 & 1.000 & 1.000 & 1.000 \\ 
        Lymphadenopathy & 0.970 & 0.950 & 0.923 & 0.977 & 0.857 \\ 
        Hiatal hernia & 1.000 & 1.000 & 1.000 & 1.000 & 1.000 \\ 
        Atherosclerotic calcification & 0.940 & 0.878 & 0.800 & 0.956 & 0.667 \\ 
        \midrule [ 0.7 pt]
        Average & 0.924 & 0.879 & 0.816 & 0.942 & 0.887 \\ 
        \bottomrule[ 1.5 pt]
    \end{tabular}
    }
\end{table}

\subsection{Discussion}

While the Abn-BLIP model demonstrates strong potential for diagnosing abnormalities from CTPA scans and generating structured radiology reports, several limitations and future directions must be acknowledged to support broader clinical applicability.

First, the framework is built on a predefined abnormality set of 32 clinically significant pulmonary and cardiovascular findings, organized under a structured, closed-set diagnostic hierarchy. This design enables high interpretability, modular querying, and precise detection of well-defined conditions. However, it inherently constrains the model's ability to generalize to rare or novel abnormalities, potentially leading to diagnostic omissions in open clinical settings. We recognize this trade-off between specificity and flexibility as a critical limitation. While the targeted design ensures clinical relevance of result and model interpretability, it reduces the open-set adaptability typically expected from general-purpose medical VLMs. Future work will explore open-set recognition and few-shot learning strategies to improve adaptability to unseen abnormalities.

Second, Abn-BLIP’s modular architecture supports extensibility across institutions and tasks. The abnormality-specific visual querying mechanism and structured reporting design can be tailored to new diagnostic contexts by integrating additional disease categories, domain-specific reporting conventions, and site-specific label taxonomies. This modularity facilitates institution-level customization while maintaining the benefits of structured report generation.

Third, the model exhibits performance variability across abnormalities due to class imbalance in the training data. Common abnormalities (e.g., pulmonary embolism, emphysema) are well represented, whereas less frequent ones (e.g., hiatal hernia, lymphadenopathy) remain underrepresented. This imbalance biases the model toward frequent classes, even with weighted loss and stratified sampling. As a result, per-class sensitivity remains uneven. Future directions include incorporating imbalance-aware training, curriculum learning, and minority-targeted data augmentation to promote equitable performance across abnormality categories.

Fourth, current evaluation relies primarily on automated metrics (e.g., BLEU, ROUGE, AUC), which, although useful, may not fully capture the clinical utility of generated reports. Radiology reporting requires coherent, context-aware narrative construction and accurate mapping between imaging findings and textual descriptions—factors that automated metrics often overlook. Incorporating expert-in-the-loop evaluation, radiologist-based quality assessment, and clinical outcome correlations will be essential for robust performance validation.

Lastly, real-world deployment will require validation across heterogeneous populations and imaging protocols. Our current datasets (BUH and INSPECT) differ in acquisition and demographics but are both U.S.-based. Broader validation including multi-center and international cohorts with diverse populations is needed to ensure fairness, generalizability, and robustness in clinical workflows.

In summary, while Abn-BLIP advances clinically structured and semantically aligned CTPA report generation, future extensions must balance structured precision with open-set flexibility, address class imbalance, and incorporate expert-driven evaluation. Broader multi-institutional validation will be critical to ensuring real-world usability, fairness, and clinical integration.

\section{Conclusion}

In conclusion, Abn-BLIP represents a significant advancement in automated medical imaging interpretation, introducing a clinically aligned vision–language framework tailored for CTPA analysis. By integrating learnable abnormality-guided queries with a hierarchical multimodal transformer (Abn-QFormer) and employing fine-grained cross-modal alignment, Abn-BLIP effectively captures abnormality-specific findings across pulmonary and cardiovascular structures.

The model demonstrates robust performance in both multi-label abnormality classification and structured radiology report generation, achieving consistent improvements in NLG and CE metrics across internal and external datasets. Expert evaluations further confirm the clinical accuracy, clarity, and relevance of the generated reports, with strong correlations observed between expert ratings and automated metrics. Qualitative visualizations and case studies highlight Abn-BLIP’s ability to localize and describe both primary and incidental findings, a critical feature for comprehensive patient management.

In addition, Abn-BLIP exhibits favorable inference efficiency with moderate computational requirements ($\sim$280M parameters), supporting its feasibility for real-world deployment. Its modular and extensible design enables adaptation across institutions and customization to local diagnostic protocols.

Overall, Abn-BLIP establishes a structured, interpretable, and clinically oriented pipeline for CTPA interpretation, marking a promising step toward trustworthy AI-assisted diagnosis and radiology workflow optimization in diverse healthcare environments.

\section*{CRediT authorship contribution statement}
Zhusi Zhong: Writing – original draft, Visualization, Validation, Software, Methodology, Investigation, Formal analysis, Data curation, Conceptualization. 
Lulu Bi, Sun Ho Ahn, Christopher J. Mullin, Michael K. Atalay, Scott Collins: Writing – review \& editing, Clinical data collection, Expert evaluation, Conceptualization. 
Yuli Wang, Zhuoqi Ma, Grayson L. Baird, Cheng Ting Lin, Webster Stayman, Todd M. Kolb, Ihab Kamel, Harrison X. Bai: Writing – review \& editing, Methodology, Conceptualization. 
Zhicheng Jiao: Supervision, Writing – review \& editing, Methodology, Conceptualization.

\section*{Declaration of competing interest}
The authors declare that they have no known competing financial interests or personal relationships that could have appeared to influence the work reported in this paper.

\section*{Data availability}
The BUH dataset is unavailable to the public due to institutional privacy restrictions. The INSPECT dataset is openly accessible for non-commercial use under a data use agreement at https://som-shahlab.github.io/inspect-website.

\section*{Acknowledgments}
The authors gratefully acknowledge Lulu Bi, Vin Somasundaram, and Shreyas Kulkarni from the Warren Alpert Medical School of Brown University for their contributions to report evaluation and annotation quality assessment.
% This research was supported by the xxx, No. xxx (to xxx.

% \bibliographystyle{elsarticle-num} 
\bibliography{refs}
% \printbibliography

\end{document}